\documentclass[10pt,conference,letterpaper]{IEEEtran}

\usepackage[letterpaper,      
            top=1in,          
            bottom=0.75in,    
            left=0.75in,      
            right=0.75in      
           ]{geometry}
\IEEEoverridecommandlockouts                             
\usepackage[T1]{fontenc}
\usepackage{color}
\usepackage{multirow}
\usepackage{makecell}
\usepackage{tabularx}
\usepackage{array} 
\usepackage{stfloats} 
\usepackage{subcaption}
\usepackage{graphicx} 
\usepackage{amsmath} 
\usepackage{amssymb}  
\usepackage{algorithmic}
\usepackage{algorithm}
\usepackage{caption}
\usepackage{booktabs}
\usepackage{siunitx}

\usepackage{enumitem}
\usepackage{threeparttable}

\def\BibTeX{{\rm B\kern-.05em{\sc i\kern-.025em b}\kern-.08em
    T\kern-.1667em\lower.7ex\hbox{E}\kern-.125emX}}

\definecolor{myGreen}{RGB}{0, 0, 0}

\usepackage[dvipsnames]{xcolor}

\title{
High and Low Resolution Tradeoffs in Roadside Multimodal Sensing
}

\author{
\IEEEauthorblockN{Shaozu Ding$^\heartsuit$}
\IEEEauthorblockA{\textit{The Polytechnic School} \\
\textit{Arizona State University}\\
Mesa, AZ, USA \\
sding32@asu.edu}
\and
\IEEEauthorblockN{Yihong Tang$^\heartsuit$}
\IEEEauthorblockA{\textit{Civil Engineering} \\
\textit{McGill University}\\
Montreal, Canada \\
yihong.tang@mail.mcgill.ca}
\and
\IEEEauthorblockN{Marco De Vincenzi}
\IEEEauthorblockA{\textit{The Polytechnic School} \\
\textit{Arizona State University}\\
Mesa, AZ, USA \\
mdevinc2@asu.edu}
\and
\IEEEauthorblockN{Dajiang Suo$^{\dagger}$}
\IEEEauthorblockA{\textit{The Polytechnic School} \\
\textit{Arizona State University}\\
Mesa, AZ, USA \\
dajiang.suo@asu.edu}
\noindent\thanks{$^\heartsuit$Equal contribution, $^{\dagger}$Corresponding author. Technical support and review was provided by the Maricopa County Department of Transportation, AZ, USA. This work was partially supported by MIT Mobility Initiative.}
}

\begin{document}

\maketitle
\thispagestyle{empty}
\pagestyle{empty}

\begin{abstract}

Balancing cost and performance is crucial when choosing high- versus low-resolution point-cloud roadside sensors. For example, LiDAR delivers dense point clouds, while 4D millimeter-wave radar, though spatially sparser, embeds velocity cues that help distinguish objects by motion and often comes at a lower price. Unfortunately, evaluating such tradeoffs in an ex-ante manner is challenging. The roadside sensor placement locations, tilt angles, types, and configuration, will influence point cloud density and distribution across the coverage area. Compounding the first challenge is the fact that different sensor mixtures often demand distinct neural network architectures to maximize their complementary strengths. Without an evaluation framework that establishes a benchmark for comparison, it is imprudent to make any claims regarding whether marginal gains actually result from higher resolution and new sensing modalities or from the algorithms. We present an ex-ante evaluation framework that tackles the two challenges above. 

First, we realized a simulation tool that builds on integer programming and can automatically iterate and compare different multimodal sensor placement strategies against sensing coverage and cost jointly. For the second challenge, inspired by human multi-sensory integration, we propose a modular deep learning-based evaluation framework to assess whether and to what extent reductions in spatial resolution can be compensated by informational richness in detecting traffic participants. Extensive experimental testing on the proposed framework shows that fusing velocity-encoded radar with low-resolution LiDAR yields marked gains (14\% AP for pedestrians and an overall mAP improvement of 1.5\% across six traffic participant classes) at the same or even lower cost than high-resolution LiDAR alone. Notably, these marked gains hold regardless of the specific deep neural modules employed in our framework. The results challenge the prevailing assumption that high-resolution solutions are always superior to low-resolution alternatives. The code and dataset will be available at \textcolor{blue}{https://github.com/ASU-Suo-Lab/Hi-Lo-Sensing}.

\end{abstract}


\section{Introduction}

\begin{figure}[!t]
    \centering
    \includegraphics[width=\linewidth,height=5cm]{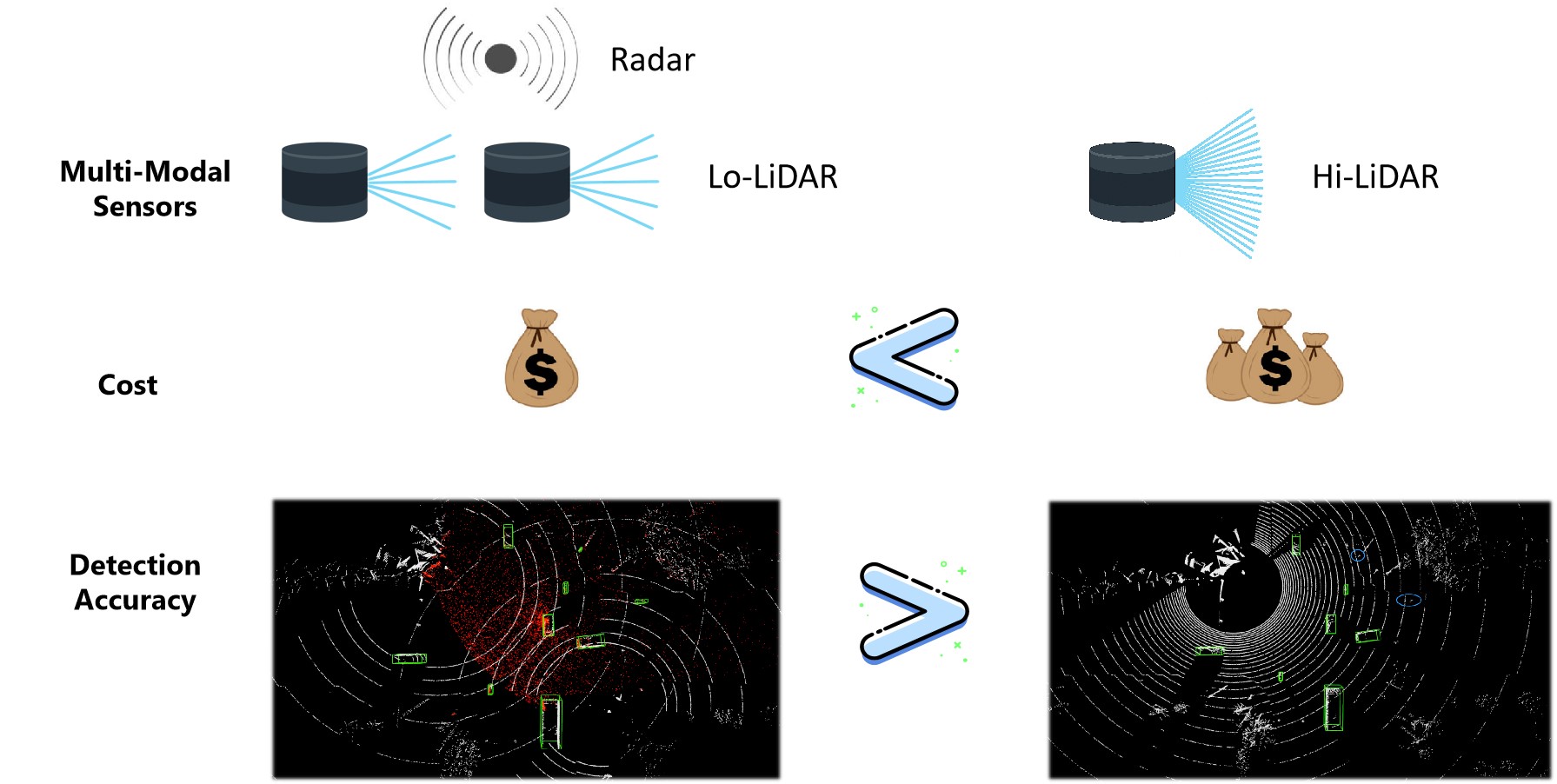}
    \caption{Cost–accuracy trade-off between a multi-modal sensing stack (two Lo-LiDARs + 4D radar) and a single baseline (Hi-LiDAR). Blue circles indicate objects that are missed by the high-resolution LiDAR alone but are successfully detected by the combination of low-resolution LiDAR and 4D radar.}
    \label{conclusion_pic}
    \vspace{-0.4cm}
\end{figure}

Infrastructure-assisted sensing mechanisms have gained significant traction in recent years as promising solutions for improving transportation safety and efficiency. Data captured by roadside sensors can not only be used for traffic monitoring and control~\cite{bai2023cyber,zhao2019detection}, but can also be shared with connected vehicles to support vehicle-infrastructure cooperative perception and driving~\cite{xu2023bridging,xiang2024v2xreal}. Previous research has explored the placement of single sensing modalities on the roadside to improve the perception of objects of interest, including LiDAR~\cite{lidarlibrary,Qu2023,jiang2023optimizing}, cameras~\cite{zou2022real}, and radar~\cite{Chen2024}, and roadside sensor in general~\cite{ma2024virtual}. However, achieving an optimal cost-performance tradeoff between high and low resolution remains challenging due to the physical heterogeneity among sensor types. For example, 4D millimeter-wave radar often generates a low-density point cloud to describe the 3D geometry of objects, compared to LiDAR, which produces a dense point cloud for each snapshot during the observation period (i.e., frame). However, radar can provide velocity-related information about objects, which, when aggregated over a longer temporal span, becomes useful in recognizing objects with varying movement patterns (e.g., fast-moving vehicles vs. slow-moving bicyclists and pedestrians). To leverage heterogeneous sensors to detect various traffic participants—each with unique shapes, sizes, and mobility levels—engineers must evaluate to what extent losses in spatial resolution in one sensor type can be compensated by the richness of information when choosing among different sensor configurations (e.g., LiDAR vs. 4D mmWave radar).

To address this problem, we designed a multimodal modular evaluation framework that simulates humans’ multi-sensory integration capabilities. As shown in Fig. \ref{conclusion_pic}, by combining two low-resolution LiDARs with a 4D radar, we obtain a multimodal configuration that costs less than a single high-resolution LiDAR, but is able to detect objects that the latter currently misses (blue circles in Fig. \ref{conclusion_pic}). Our simulation experiments based on real-world intersections show that fusing low-resolution LiDAR with velocity-coded radar improves pedestrian recognition accuracy by 14\% on average and 1.5\% on average overall, at a cost roughly equivalent to that of a single high-resolution lidar, and these improvements are consistent across different neural network architectures. Therefore, our framework provides clear, data-driven guidelines for multimodal fusion strategies: a velocity-rich 4D radar can compensate for the reduced resolution of LiDAR, enabling economical yet powerful infrastructure-based perception.

In addition to perception, we also work on actual placement. We build an integer programming model to optimize decisions about sensor type, location, and number subject to budget constraints, allowing for ex-ante evaluation of sensor installations before any field work begins. Our key contributions are summarized as follows.

\begin{itemize}[leftmargin=*]
\item Multimodal low-cost and high-performance perception framework. A modular deep learning evaluation framework inspired by human multi-sensory fusion is proposed to systematically quantify the cost-performance trade-off of "low-resolution LiDAR + 4D radar" versus single high-resolution LiDAR. Experiments show that under the premise of comparable hardware expenditure, this multimodal combination can improve the detection accuracy of valuable road users, verifying the ability of velocity information to compensate for the lack of spatial resolution.
\item Simulation-driven sensor deployment optimization platform: We build an integer programming model and digital twin simulation platform to perform ex-ante planning for multimodal sensor types, quantities and locations. The platform can optimize field of view coverage under cost constraints and provide a common comparison benchmark for multimodal perception  of roadside infrastructure.

\end{itemize}

\section{Related Work}

Most cooperative-perception studies concentrate on algorithmic advances (e.g., object detection networks and vehicle-to-infrastructure fusion pipelines) while the spatial optimization of roadside sensing units remains under-explored, especially for multimodal installations. Cai et al. \cite{lidarlibrary} introduced the Realistic LiDAR Simulation (RLS) library, which replicates the proprietary characteristics of several commercial LiDARs within CARLA and demonstrates that LiDAR placement has a significant impact on perception accuracy. Jiang et al. \cite{jiang2023optimizing} further released a roadside synthetic benchmark dedicated to placement research and proposed a greedy search that consistently outperforms empirical or uniform placements under an equal-sensor budget. Complementing these LiDAR-only efforts, Vijay et al. \cite{Vijay} formulated placement as a cost-coverage-redundancy trade-off, coupling ray casting visibility analysis with linear programming to yield near-optimal solutions that increase critical zone coverage within fixed budgets. All aforementioned methods treat the infrastructure as a single-modality network; cross-modal synergies among LiDAR, radar grids are neither modeled nor optimized. Consequently, a principled framework for the joint placement of multimodal sensors is still missing from the existing research.

3D object detection is an important part of in-vehicle or roadside perception systems. Models like PointPillars \cite{lang2019pointpillars} and CenterPoint \cite{yin2021centerbased3dobjectdetection} are widely used for efficient road user detection. The recent trend explores the integration of 4D millimeter-wave radar with complementary sensors, leveraging the strengths of each modality to offset individual weaknesses and achieve higher perception accuracy. RadarNet \cite{yang2020radarnetexploitingradarrobust} is the first algorithm to propose the fusion of LiDAR and 3D radar modalities. It effectively overcomes radar noise and ambiguity through early voxel fusion and late attention fusion, and has achieved good results in target detection and velocity estimation tasks. Bi-LRfusion \cite{wang2023bilrfusionbidirectionallidarradarfusion} proposes a bidirectional LiDAR and 3D radar fusion framework, allowing the radar branch to learn local details from the LiDAR branch, alleviating the problem of height loss and sparsity. It can significantly improve the accuracy of dynamic target detection while maintaining lightweight, and achieve good results on the nuScenes and ORR datasets. LiRafusion \cite{song2024lirafusiondeepadaptivelidarradar} proposes a joint voxel feature encoder and an adaptive gating network to enhance the complementary information of LiDAR and 3D radar, fuse cross-modal feature maps, and achieve significant performance improvement on the nuScenes dataset. L4DR \cite{huang2025l4drlidar4dradarfusionweatherrobust} pioneered an early complementary fusion framework for LiDAR and 4D radar. It solved the problem of poor single-modality perception in bad weather by fusion of foreground denoising and parallel gated features, and achieved significant detection gains on VoD and K-Radar datasets.

Although these advanced studies have discussed the multimodal fusion framework of LiDAR and 3D or 4D radar in depth, they are all based on the perspective of single sensor resolution, and different modal combinations often require tailored network architectures to complement each other. No framework has considered the trade-off between the cost and performance of multimodal sensors. Therefore, based on a real roadside test bench, this article generates perception data of different resolutions through simulation driving, compares multiple LiDAR and radar fusion strategies, and proposes the first modular deep learning evaluation framework that considers the relationship between spatial resolution and information richness.

\begin{algorithm}[thbp]
    \footnotesize
    \caption{Multi-modal Sensor placement}
    \label{EmbeddedAlg}
    \renewcommand{\algorithmicrequire}{\textbf{Input:}}
    \renewcommand{\algorithmicensure}{\textbf{Output:}}
    \begin{algorithmic}[1]
        \footnotesize
        \REQUIRE 
        \parbox[t]{\linewidth}
        {Visibility of LiDAR $V^l$ and radar $V^r$.}
        \ENSURE 
        \parbox[t]{\linewidth}
        {placement points of LiDAR $S_l$ and radar $S_r$.}
        \scalebox{0.9}{
        \begin{minipage}{\linewidth}
        \begin{align}
        \max \quad & \sum_{j}^{N_T} t_j \cdot \rho_j, & 
        \label{eq_target}\\
        \mbox{s.t.}\quad
        &\sum_{i}^{N_l} x_i + \sum_{i}^{N_r} y_i \leq N
        \label{eq_numlimit}\\
        &\sum_{i}^{N_l}-\ln (1-V^l_{(i,j)}) \cdot x_i \geq t_j, \; \forall j \in 1,...,N_T 
        \label{eq_Lvisable}\\
        &\sum_{i}^{N_r}-\ln (1-V^r_{(i,j)}) \cdot y_i \geq t_j, \; \forall j \in 1,...,N_T 
        \label{eq_Rvisable}\\
        &\rho_j = (\sum_{i}^{N_l}V^l_{(i,j)}x_i + \sum_{i}^{N_r} V^r_{(i,j)}y_i)\cdot w_j, \;
        \forall j \in 1,...,N_T
        \label{eq_average}\\
        &x_i \in \{0,1\}, \; \forall i \in 1,...,N_l 
        \label{eq_lidar}\\
        &y_i \in \{0,1\}, \; \forall i \in 1,...,N_r  
        \label{eq_radar}\\
        &t_j \in \{0,1\}, \; \forall j \in 1,...,N_T  
        \label{eq_t}
        \end{align}
        \end{minipage}
        }
    \end{algorithmic}
\end{algorithm}

\section{Multimodal Sensor Placement Optimization}

Before evaluating different roadside sensing systems, each utilizing different combinations of sensing modalities and resolutions, we need to determine the optimal deployment position for each sensor. The purpose is to provide researchers with a common comparison benchmark. Unlike vehicle-mounted sensing, roadside sensing avoids comparisons based on different benchmarks due to the uncertainty of the location and posture of the deployed sensors.

\begin{figure*}[!t]
    \centering
    \includegraphics[width=.9\textwidth]{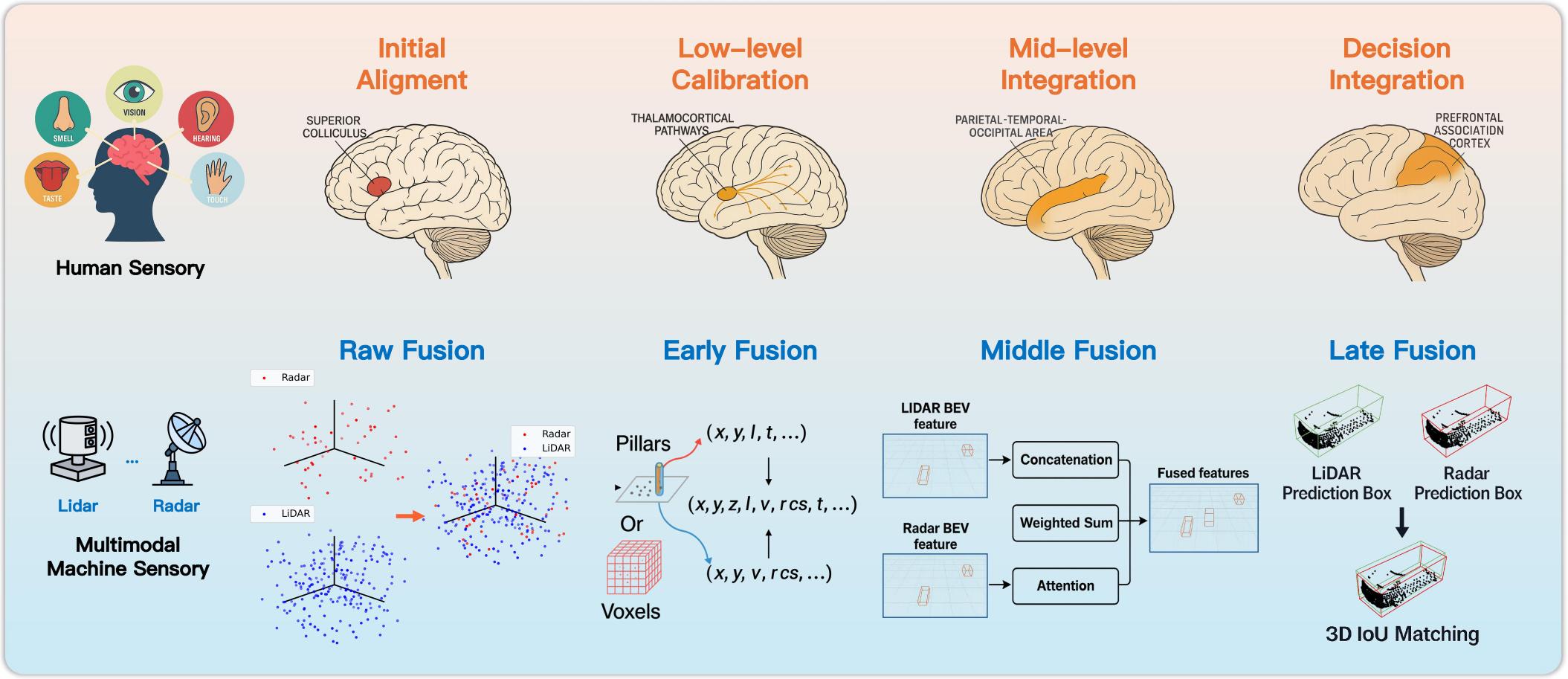}
    \caption{An analogy between human sensory integration and multimodal machine perception.}
    \label{fig:method}
    \vspace{-0.5cm}
\end{figure*}
Building on previous work \cite{lidarlibrary, Qu2023, Vijay} on the implementation of LiDAR placement, we define sensor visibility, allowing us to quantify the coverage area of each sensor. Formally, we divide the road network into a grid to facilitate sensor placement and define the region of interest (ROI) as \( R = \bigcup R_i \), where each \( R_i \) denotes an individual road segment. To handle different sensor modalities, we define two sets of candidate points: one for LiDAR, \( S^l = \{s_1^l, s_2^l, \dots, s_{N_l}^l\} \), and another for radar, \( S^r = \{s_1^r, s_2^r, \dots, s_{N_r}^r\} \), where each \( s_i^l \) and \( s_i^r \) specifies a possible LiDAR and radar location, respectively. The target points, which represent specific grid cells within the ROI that require monitoring, are denoted by \( T = \{t_1, t_2, \dots, t_{N_T}\} \subset R \). Sensor visibility is captured using separate binary visibility matrices for each modality: \( V^l \in \mathbb{R}^{N_l \times N_T} \) for LiDAR and \( V^r \in \mathbb{R}^{N_r \times N_T} \) for radar. Here, \( V_{i,j}^l = 1 \) or \( V_{i,j}^r = 1 \) indicates that the respective LiDAR or radar sensor \( s_i \) can detect the target point \( t_j \), and \( V_{i,j}^l = 0 \) or \( V_{i,j}^r = 0 \) otherwise. 

We propose Algorithm \ref{EmbeddedAlg} based on integer programming (IP) to optimize the placement of multimodal road sensors under a budget constraint, which extends the previous study of roadside LiDAR placement \cite{Qu2023}. The goal of Algorithm \ref{EmbeddedAlg} is to maximize the visible area as described in \eqref{eq_target}, where the sensor visibility conditions (expressed in \eqref{eq_Lvisable} and \eqref{eq_Rvisable} mean that a given target grid is considered visible only if it is “seen” by both LiDAR and radar. The weight matrix $\rho_j$ contains the sensor visibility probability and the importance of each region or grid, which is calculated as \eqref{eq_average}. For example, if the goal of traffic monitoring is to detect potential conflicts between vehicles and vulnerable road users (VRUs) in the central area, a higher weight can be assigned to this area than to the peripheral oncoming lanes connecting the intersection. The binary variables $x_i$ and $y_i$ represent the candidate points for LiDAR and radar sensors, respectively. Equation \eqref{eq_numlimit} imposes a limit on the total number of sensors, which is a simplified form of the cost constraint.

\begin{figure*}[htbp]
    \centering
    \includegraphics[width=.9\textwidth]{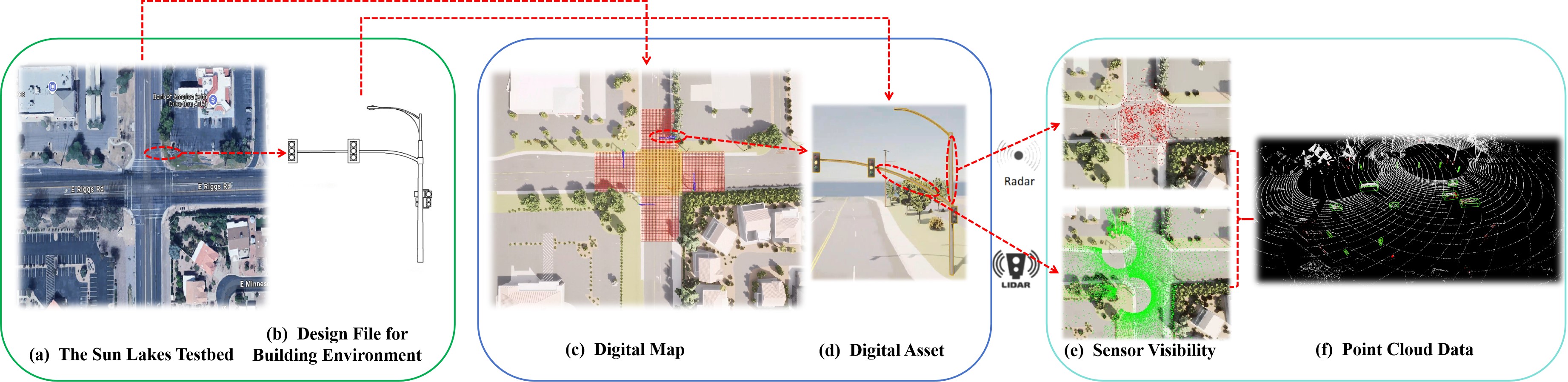}
    \caption{Illustrating the ex-ante evaluation process through a case study on the Sun Lakes Test Bed in Sun Lakes, AZ, USA.}
    \label{digitalmap}
    \vspace{-0.5cm}
\end{figure*}

\section{Biologically Inspired Multimodal Fusion Framework for Roadside Sensing}
Building on optimized sensor placements described previously, we propose a biologically inspired multimodal fusion framework to evaluate roadside perception capabilities by different sensor configurations. Our framework takes inspiration from human sensory integration, particularly the hierarchical integration processes in the brain. Human perception progressively integrates sensory inputs, such as vision, sound, and motion, through increasingly sophisticated levels of abstraction, ultimately enabling complex decision-making. Similarly, our multimodal perception approach comprises four progressive fusion stages, raw fusion, early fusion, middle fusion, and late fusion, as depicted in Fig.~\ref{fig:method}. This structured approach systematically exploits geometric and semantic complementarities across different sensing modalities, incrementally transforming raw multimodal data into actionable insights.

Formally, we represent the input LiDAR and radar point clouds as \(\mathcal{P}^L = \{p^L_i\}\) and \(\mathcal{P}^R = \{p^R_j\}\), respectively. Each point from these modalities is mapped into a common \(d\)-dimensional feature space using modality-specific encoders \(f_L\) and \(f_R\). Thus, the encoded features are defined as \(f(p) = f_L(p)\) if \(p \in \mathcal{P}^L\) and \(f(p) = f_R(p)\) if \(p \in \mathcal{P}^R\).

In the Raw Fusion stage~\cite{lang2019pointpillars,yin2021centerbased3dobjectdetection}, analogous to the initial alignment performed by the superior colliculus in the human brain~\cite{meredith1983interactions}, we directly combine the point-level features from both modalities without imposing additional structure. The resulting fused point-wise feature set \(\mathcal{F} = \{f(p)\ |\ p \in \mathcal{P}^L \cup \mathcal{P}^R\}\) is input into a shared backbone network for subsequent processing.

During the Early Fusion stage~\cite{yang2020radarnetexploitingradarrobust,song2024lirafusiondeepadaptivelidarradar,huang2025l4drlidar4dradarfusionweatherrobust}, similar to the low-level calibration processes in the human thalamocortical pathways~\cite{cappe2009multisensory}, the features from each modality are first structured into 3D spatial partitions. Specifically, we voxelize or pillarize the combined space, aggregating features within each grid cell \(g \in \mathcal{G}\), where \(\mathcal{G}\) denotes the set of non-empty grid cells. The aggregated features are computed as \(F^{\text{early}}_g = A(\mathcal{F}_{g})\), with aggregation function \(A\) typically being mean or max pooling. The output is a BEV-aligned feature map \(F^{\text{early}}\).

In the Middle Fusion stage~\cite{wang2023bilrfusionbidirectionallidarradarfusion,song2024lirafusiondeepadaptivelidarradar,huang2025l4drlidar4dradarfusionweatherrobust}, paralleling cortical integration within the parietal-temporal-occipital association areas~\cite{beauchamp2004unraveling}, each modality is first individually encoded into grid-aligned representations \(F^L\) and \(F^R\). These representations are subsequently merged using a fusion operator \(\phi\). Common fusion methods include concatenation \((\phi(F^L, F^R) = [F^L; F^R])\), weighted sums \((\phi(F^L, F^R) = \alpha F^L + (1 - \alpha) F^R)\), or attention mechanisms \((\phi(F^L, F^R) = \text{Attn}(F^L, F^R))\). The fused representation \(F^{\text{mid}}\) is then utilized by the downstream detection network.

Finally, the Late Fusion stage mimics decision-level integration occurring in the prefrontal cortex~\cite{romanski2007representation}. At this stage, each modality independently generates a set of detection boxes, denoted as $\mathcal{D}^L$ from LiDAR and $\mathcal{D}^R$ from radar. These are then fused through a geometry-based decision fusion operator $\Psi$, which performs 3D IoU matching to align and merge prediction boxes from both modalities. Specifically, detection pairs with high 3D overlap are matched, and their attributes (e.g., location, size, score) are refined or averaged based on agreement. This process enhances robustness by leveraging the spatial precision of LiDAR and the velocity cues from radar. The final detection set is expressed as $\mathcal{D} = \Psi_{\text{IoU}}(\mathcal{D}^L, \mathcal{D}^R)$.

This biologically inspired framework provides a structured approach to systematically investigate multimodal fusion strategies, sensor combinations, and resolution trade-offs, directly supporting the case study in subsequent sections.

\section{High and Low-resolution tradeoff exploration through a real-world case Study}

The tradeoff between high and low-resolution sensing is investigated through a case study at an intersection testbed in Sun Lakes, Chandler, Arizona, USA. Due to the resort-like nature and numerous country clubs in Sun Lakes, this area experiences higher volumes of vulnerable road users (VRUs). The local Department of Transportation (DOT) seeks to improve its situational awareness to detect potential conflicts between vehicles and VRUs, to enhance safety. This motivates our exploration of cost-effective sensing solutions that can be scaled to other intersections with similar needs.

To minimize the costs of sensor relocation or reconfiguration, we utilize a simulation-assisted framework for ex-ante evaluation of sensor placement strategies with varying modalities and resolutions (Fig. \ref{digitalmap}). A realistic digital replica of the Sun Lakes testbed was built using publicly available data and input from stakeholders, including the local DOT.
\begin{figure*}[thbp]
        \centering
        \subfloat[Central coverage = 32.4\%]{
        \includegraphics[width=0.3\linewidth,height=3.5cm]{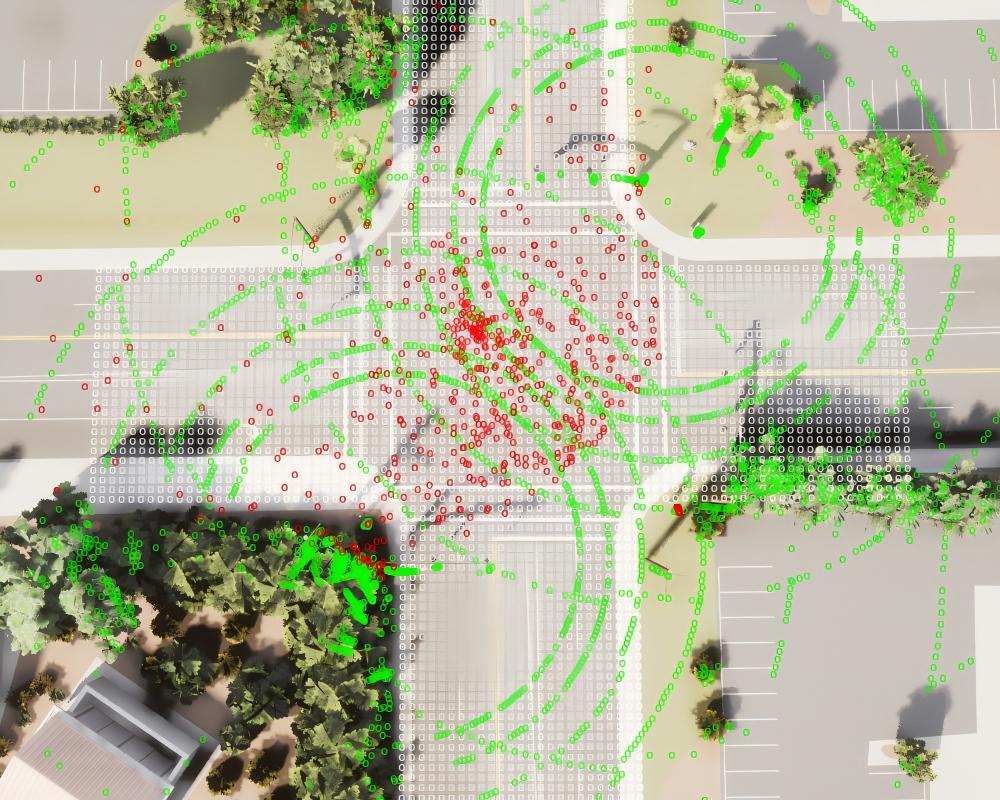}
        }
        \subfloat[Central coverage = 34.5\%]{
       \includegraphics[width=0.3\linewidth,height=3.5cm]{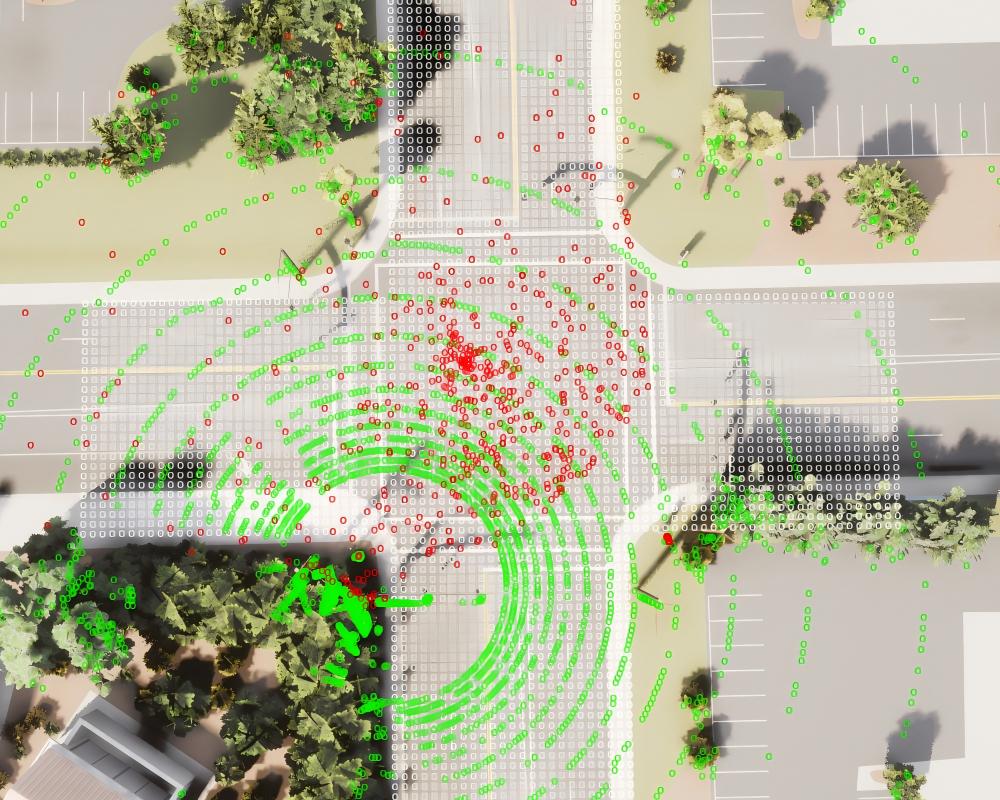}
        }
        \subfloat[Central coverage = 41.4\%]{
        \includegraphics[width=0.3\linewidth,height=3.5cm]{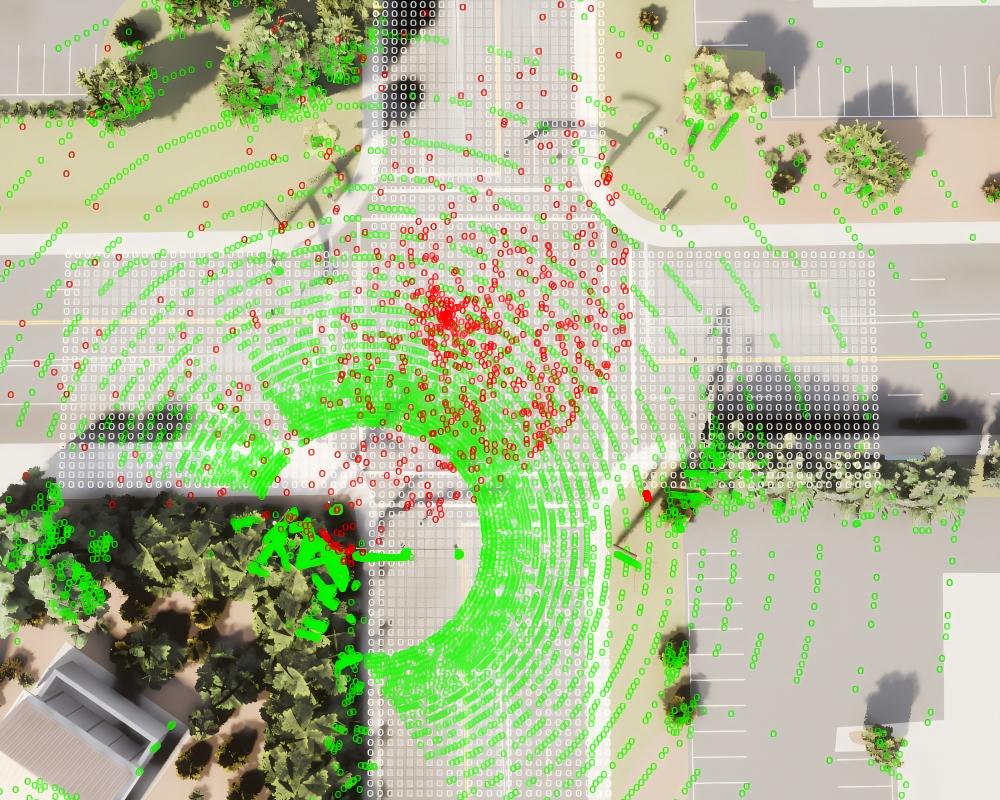}
        }
        \caption{An example of roadside multi-modal sensor placement optimization. (a) is the placement of two 16-beam LiDARs and one 4D millimeter-wave radar after optimization. (b) is the optimized placement of one 32-beam LiDAR and one 4D radar. (c) is the optimized placement of one 64-beam LiDAR and one 4D radar. The green and red dots denote the LiDAR and radar point clouds respectively.}
\label{coverage_vis}
\vspace{-0.4cm}
\end{figure*}
\begin{itemize}[leftmargin=*]
    \item Digital maps of the intersection: We used GIS and aerial imagery to create an accurate map of the intersection, incorporating the road network and static objects (e.g., buildings, trees) using RoadRunner, a tool for 3D scene editing in autonomous vehicle applications.
    \item Region of interest (ROI): The central region (yellow in Fig. \ref{digitalmap}(c)) focuses on detecting vehicle and VRU conflicts. For different applications, such as adaptive traffic signal, the ROI may cover vehicle queues (red in Fig. \ref{digitalmap}(c)).
    \item Potential sensor placement locations: They were determined in consultation with the local DOT managing the intersection. While we are authorized to place sensors on traffic light poles at the intersection’s four corners, practitioners may select alternative locations to optimize visibility. In such cases, Algorithm \ref{EmbeddedAlg}’s sensor limit, based on procurement costs, can be adjusted to account for construction and maintenance costs (e.g., new poles, power, and communication lines).
    \item Derivation of traffic and VRU flows for simulation: Traffic engineers provided reasonable estimates of vehicle and VRU flows for the simulation. Alternative methods, such as using vehicle-generated movement data \cite{2019Zhan} or drone-captured aerial footage \cite{Zheng_2023}, can also be applied.
\end{itemize}

\begin{table}[!b]
  \caption{LiDAR and 4D Radar specifications}
  \label{tab:sensor_specs}
  \centering
  \begin{tabularx}{\linewidth}{@{}l c X@{}}
    \toprule
    \textbf{Sensor Type} & \textbf{Resolution} & \textbf{Details}   \\ 
    \midrule
    \multirow[c]{3}{*}{LiDAR}
       & High & 64\,beams, 20\,Hz, 360°\,HFOV, 45°\,VFOV, $\le$90\,m \\
       & Mid  & 32\,beams, 20\,Hz, 360°\,HFOV, 45°\,VFOV, $\le$90\,m \\
       & Low  & 16\,beams, 20\,Hz, 360°\,HFOV, 30°\,VFOV, $\le$100\,m \\ 
    \midrule
    Radar & High & 20\,Hz, 120°\,HFOV, 28°\,VFOV, $\le$90\,m  \\
       \bottomrule
  \end{tabularx}
  \vspace{-0.
  cm}
\end{table}

\section{Experimental results and Discussion}
\subsection{Experimental Settings.}
We conduct the sensor-placement study in CARLA \cite{dosovitskiy2017carla}, sampling 590 candidate mounting points along the traffic light poles at the four corners of the Sun Lakes testbed. The sensor options - which cover multiple modalities and resolutions - are listed in Table \ref{tab:sensor_specs}. Placement is optimized with Algorithm \ref{EmbeddedAlg}, solved as a mixed-integer program in Gurobi.

We generate a multi-agent roadside dataset in CARLA by mounting different LiDAR and radar combinations with high and low resolution on the traffic poles of the simulated intersection. The data are organised similarly as in nuScenes\cite{nuscenesdataset}, allowing immediate use with existing pipelines. In total, 37 traffic flow scenes of 20 s each are captured at 20 Hz, yielding 15,000 frames (750 s). We adopt a 80\% / 20\% train / test split. Across these frames, we annotate 135,333 bounding boxes spanning six classes: car, truck, motorcycle, bus, pedestrian, and golf cart, the last of which is rarely covered in public datasets and is therefore newly modeled in UE4 for this work. To our knowledge, this is the first large-scale benchmark with multi-resolution and multimodal sensors for roadside perception, allowing fair comparison of multi-modal detection and fusion methods.

\subsection{Multi-modal Sensor Placement Optimization}
In order to quantify the sensor's field of view, we define the proportion of grids in the ROI area that are passed by ray casting as the central coverage area. Fig. \ref{coverage_vis} shows the central coverage of the three combinations of multimodal sensors after optimization of placement. The distribution of radar point clouds in the three cases is similar, indicating that the placement algorithm has fixed the radar in the advantageous position at the center of the intersection. As the LiDAR resolution increases, the overlap area of LiDAR and radar expands, providing more complementary features for fusion detection and improving redundancy. By comparing (a) with (b) and (c), we can find that higher-resolution LiDAR can achieve higher center coverage than multiple low-resolution ones, but it is worth noting that it can only prove that the coverage area increases monotonically with the improvement of LiDAR resolution, and cannot prove that there is positive feedback on the detection accuracy of valuable road users.

Emphasizing roadside perception placement algorithms gives the research community a unified, repeatable geometric benchmark. Unlike onboard sensors that are fixed in the vehicle frame as the scene moves, roadside sensors are anchored in the world frame. Without a standardized placement procedure and published sensor poses, mAP and other metrics are measured in incomparable setups and cannot reveal true model quality. By casting placement as a constrained optimization problem and releasing both the sensor configuration and optimized poses, we eliminate this variability. Researchers can then evaluate LiDAR-only, radar-only, and multimodal-fusion methods on the same coverage-versus-cost baseline, ensuring that algorithm improvements, dataset extensions, and engineering placements are all judged within a fair and consistent framework.

\begin{figure}[!t]
    \centering
    \subfloat[Per-algorithm mean average precision (mAP)]{
        \includegraphics[width=\linewidth,height=3.8cm]
        {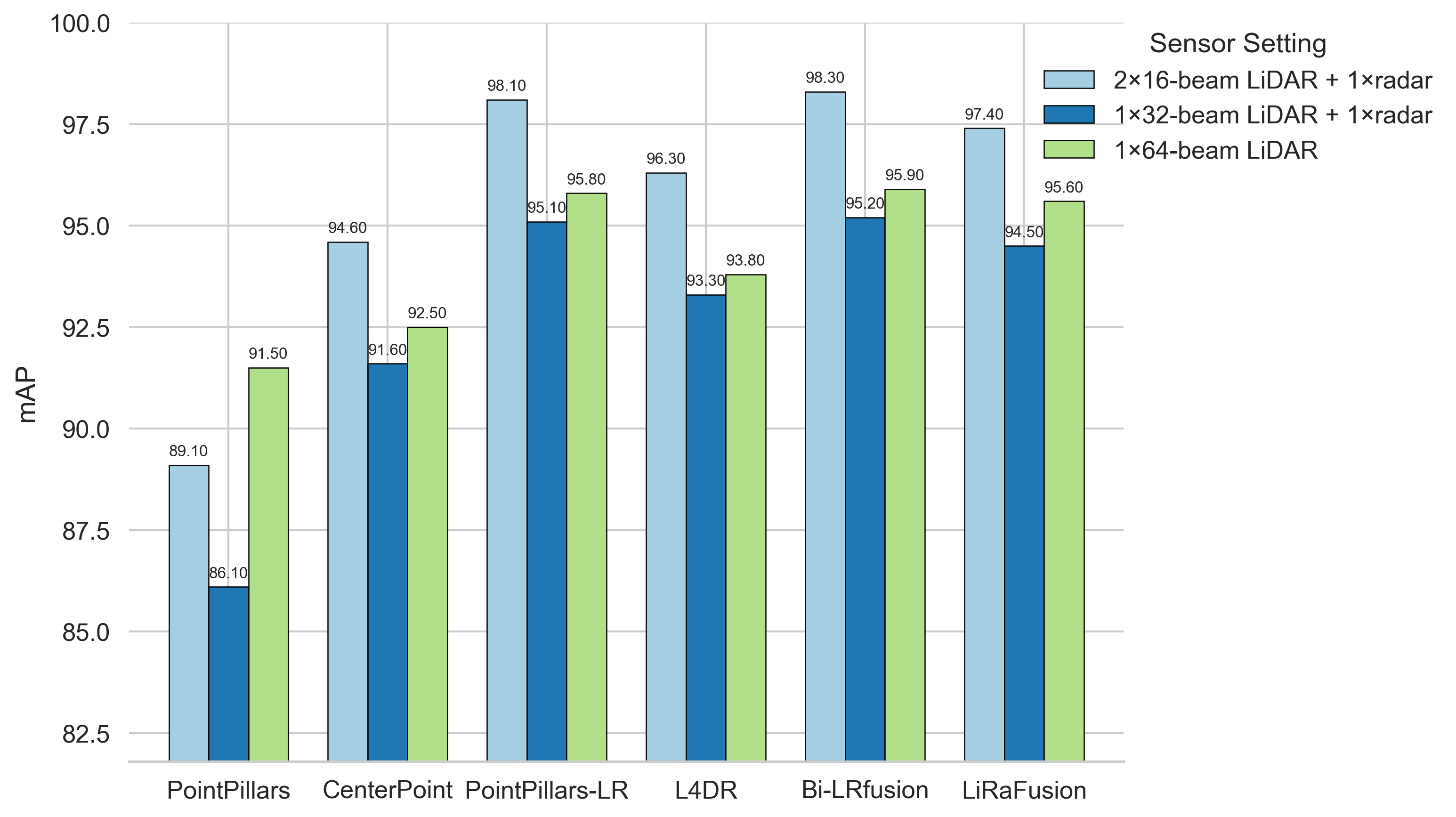}
        }
        
    \subfloat[Per-algorithm pedestrian average precision (AP)]{
       \includegraphics[width= \linewidth,height=3.8cm]
       {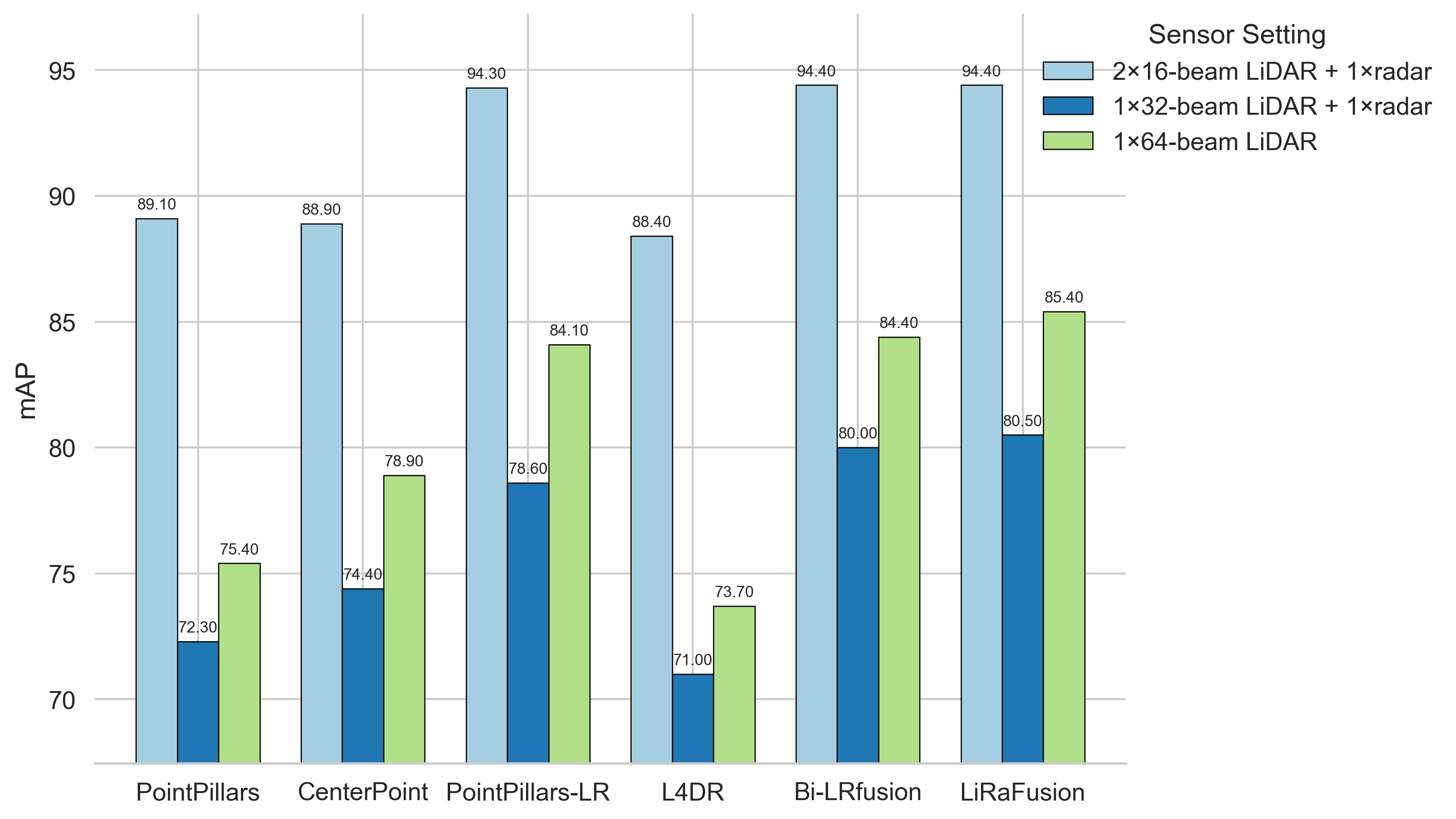}
        }
    \caption{Per-algorithm performance under three LiDAR resolutions with the same 4D Radar configuration. PointPillars-LR represents our improvement of PointPillars that introduces 4D Radar branch for fusion.}
    \label{mAP_comparison}
    \vspace{-0.4cm}
\end{figure}

\subsection{Multimodal Fusion Framework for Roadside Sensing}

We assess the cost-performance tradeoffs by comparing the performance achieved by different single-modal sensor perception algorithms and multi-modal sensor fusion perception algorithms. We train two classic LiDAR-based and four mainstream LiDAR and radar fusion-based neural network architectures on our dataset. The test results are shown in the Fig. \ref{mAP_comparison}. As shown in Fig. \ref{mAP_comparison}(a), the average mAP of four architectures which introduce the radar modality branch is 7.7\% and 3.1\% higher than that of PointPillars and CenterPoint architectures based on only LiDAR architectures, respectively, and the gap gradually narrows as the LiDAR resolution increases. It indicates that the doppler velocity, radar cross-sectional area (RCS) and other information provided by the radar can effectively make up for the sparse, occluded or insufficient geometric details of the low-resolution LiDAR point clouds, thereby significantly enhancing the detection performance. The results verify that multimodal fusion can achieve or exceed the perception accuracy of high-resolution LiDAR at a low-cost configuration. In addition, the mAP of any perception algorithm under two 16-beam LiDARs and one radar is always higher than the mAP of other high-resolution combinations of the same algorithm, which is improved by 3.2\% and 1.5\% compared to 32- and 64-beam, while the cost is reduced by 26.7\% and 56\%, respectively. The results give us inspiration that, under certain budget constraints, considering low-resolution LiDAR and fusing multimodal information is more cost-effective and has better performance than simply improving the LiDAR resolution.

\begin{figure}[!t]
    \centering
    \includegraphics[width=0.9\linewidth]{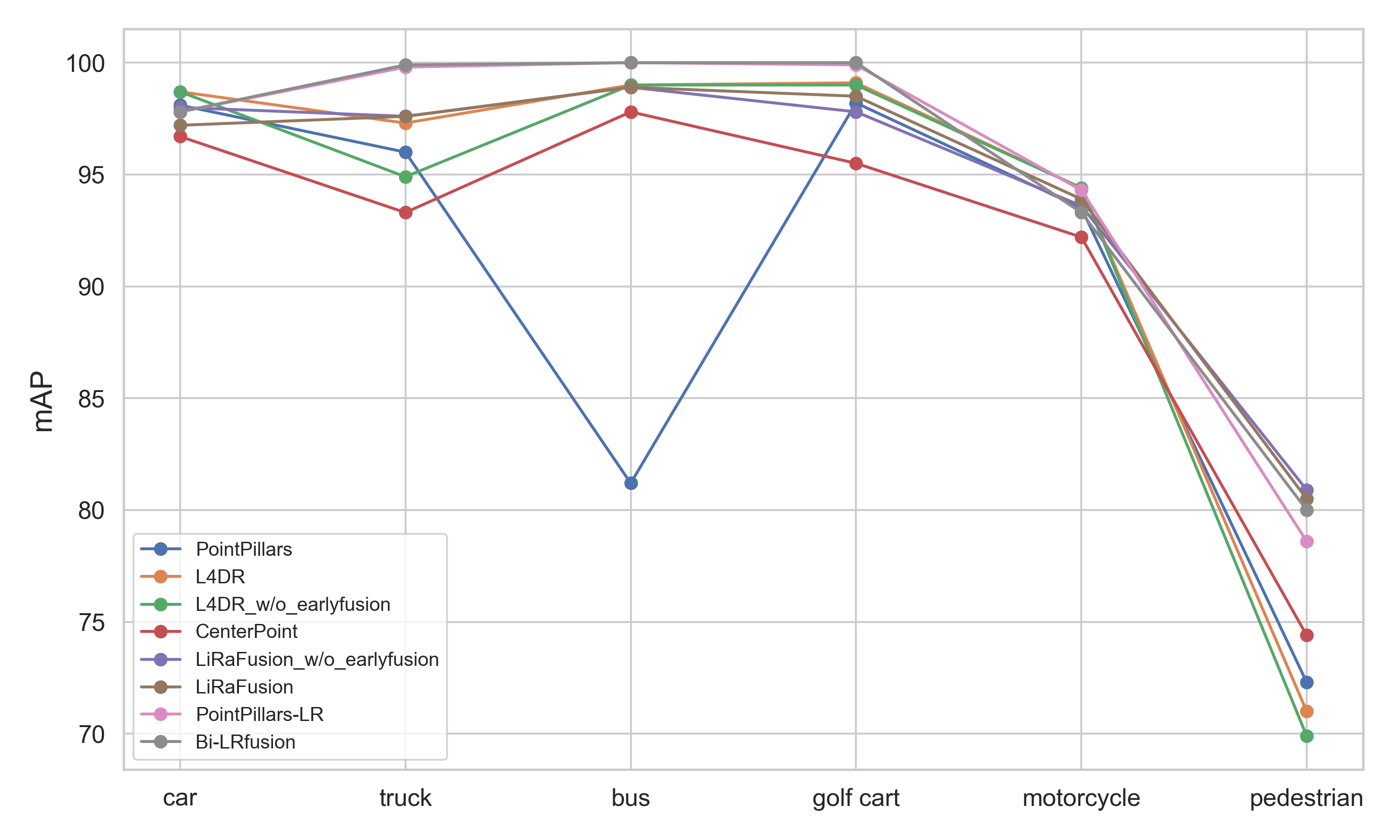}
    \caption{Per-Class AP Comparison for Detection Algorithms.}
    \label{AP_1x64_comparison}
    \vspace{-0.3cm}
\end{figure}

Focusing on pedestrians, the most vulnerable road users, we observe that augmenting LiDAR with radar info modality boosts pedestrian AP by 9.2\% over the LiDAR-only PointPillars baseline and by 6.8\% over CenterPoint, excluding L4DR architecture, underscoring the value of multi-modal fusion for fine-grained detection. At the same time, complementary low-resolution LiDAR and radar that introduce speed information can effectively improve pedestrian detection accuracy. Compared to a single high-resolution LiDAR, mAP increased by 20.3\% and 14\% respectively, indicating that the trade-off between spatial resolution and information richness is particularly significant for pedestrian detection. We also conduct ablation experiments on two multimodal architectures that include early fusion. As shown in Fig. \ref{AP_1x64_comparison}, the results show that removing early fusion will slightly reduce the detection accuracy, with a decrease of 0.7\% and 0.9\% for L4DR and LiRaFusion respectively. This verifies that the initial combination of the original information of two sensors in the early fusion stage can achieve multimodal performance gains without increasing hardware costs.

\section{CONCLUSIONS}


In this work we presented a framework for roadside multimodal sensing that balances cost and detection performance through integer programming based sensor placement and a biologically inspired fusion pipeline. Our simulations of an Arizona intersection show that fusing two low resolution LiDARs with one 4D radar at cost parity with a single high resolution LiDAR yields a fourteen percent increase in pedestrian detection and overall mean average precision gains across various fusion networks. A detailed comparison of the visual detection results are shown in Fig. \ref{fig:vis_pred}. By publishing optimized sensor layouts and an accompanying benchmark dataset we provide the community with a standard for fair evaluation of cooperative perception methods. Future steps include incorporating additional sensor types, such as thermal cameras, modeling dynamic environmental factors, and conducting real-world deployments to validate and refine our approach to scalable infrastructure sensing that improves roadway safety.

\clearpage

\newcolumntype{C}[1]{>{\centering\arraybackslash}m{#1}}
\begin{figure*}[htbp]
  \centering
  \vspace{-1cm}
  \setlength\tabcolsep{1pt}
  \begin{tabular}{C{6pt} C{0.33\linewidth} C{0.33\linewidth} C{0.33\linewidth}}
     & \textbf{2×16-beam LiDARs + 1×4D Radar} & \textbf{1×32-beam LiDAR + 1×4D Radar} & \textbf{1×64-beam LiDAR} \\[6pt]

     \rotatebox{90}{Pointpillars} &
      \includegraphics[width=0.3\textwidth,height=3.3cm]{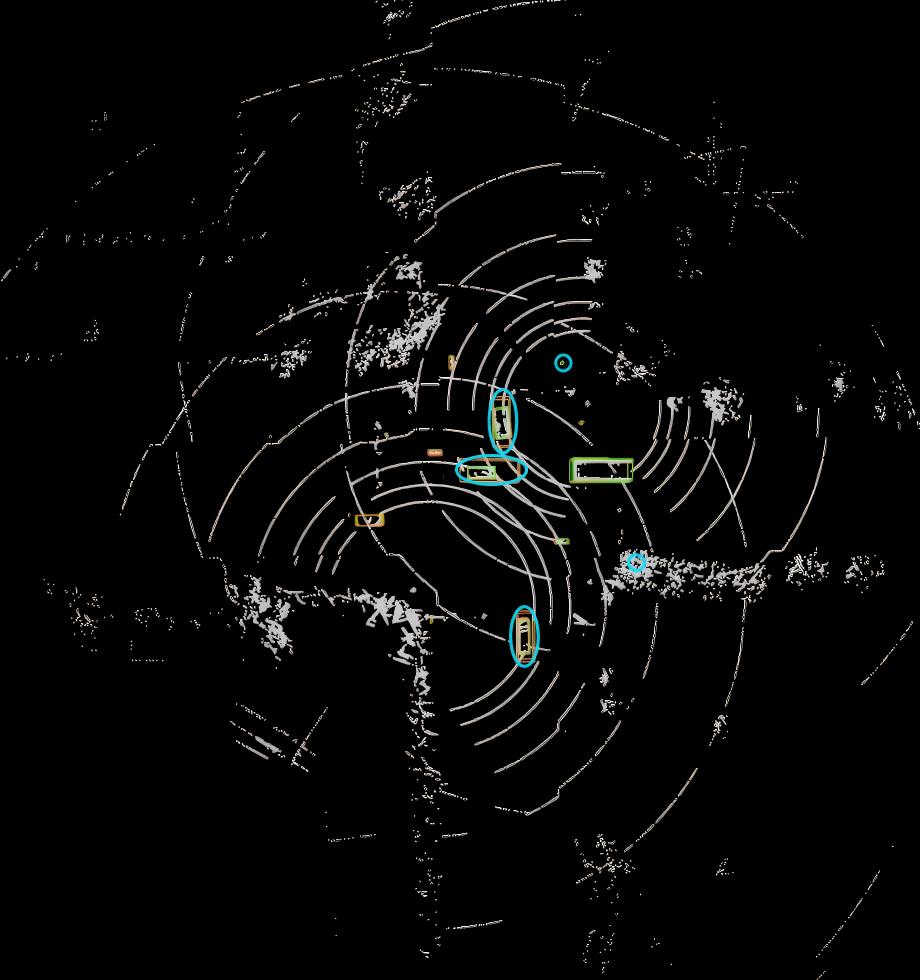} &
      \includegraphics[width=0.3\textwidth,height=3.3cm]{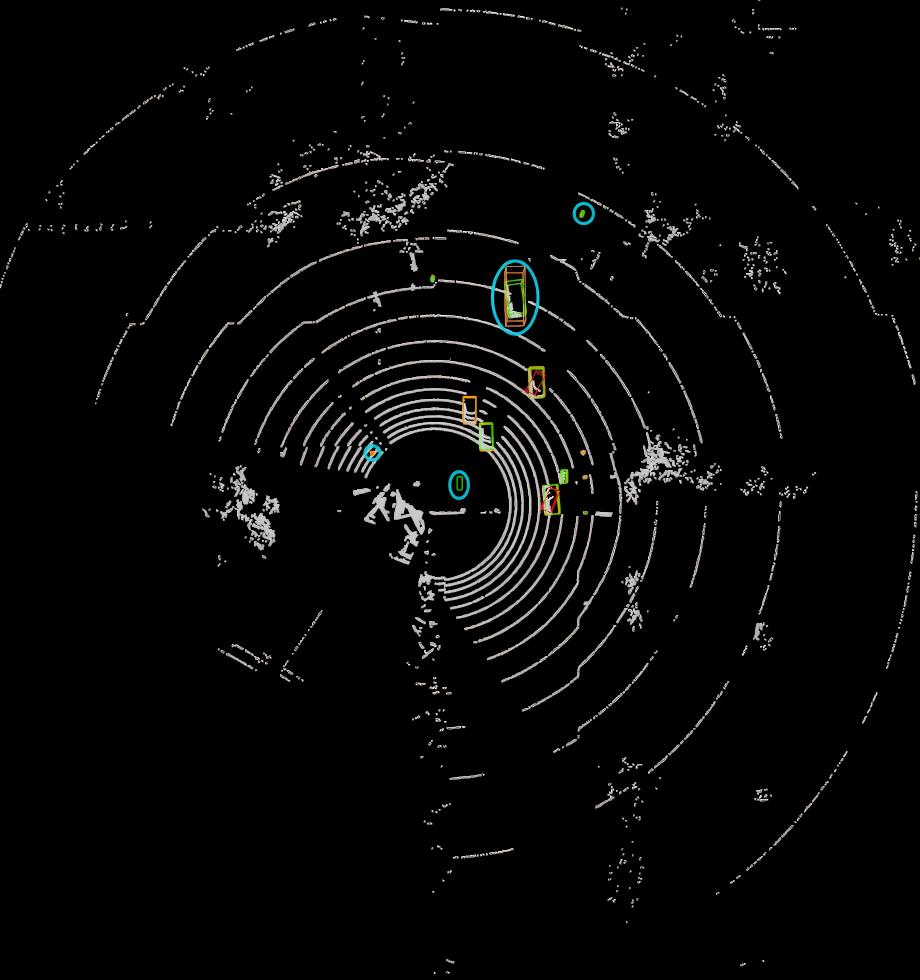} &
      \includegraphics[width=0.3\textwidth,height=3.3cm]{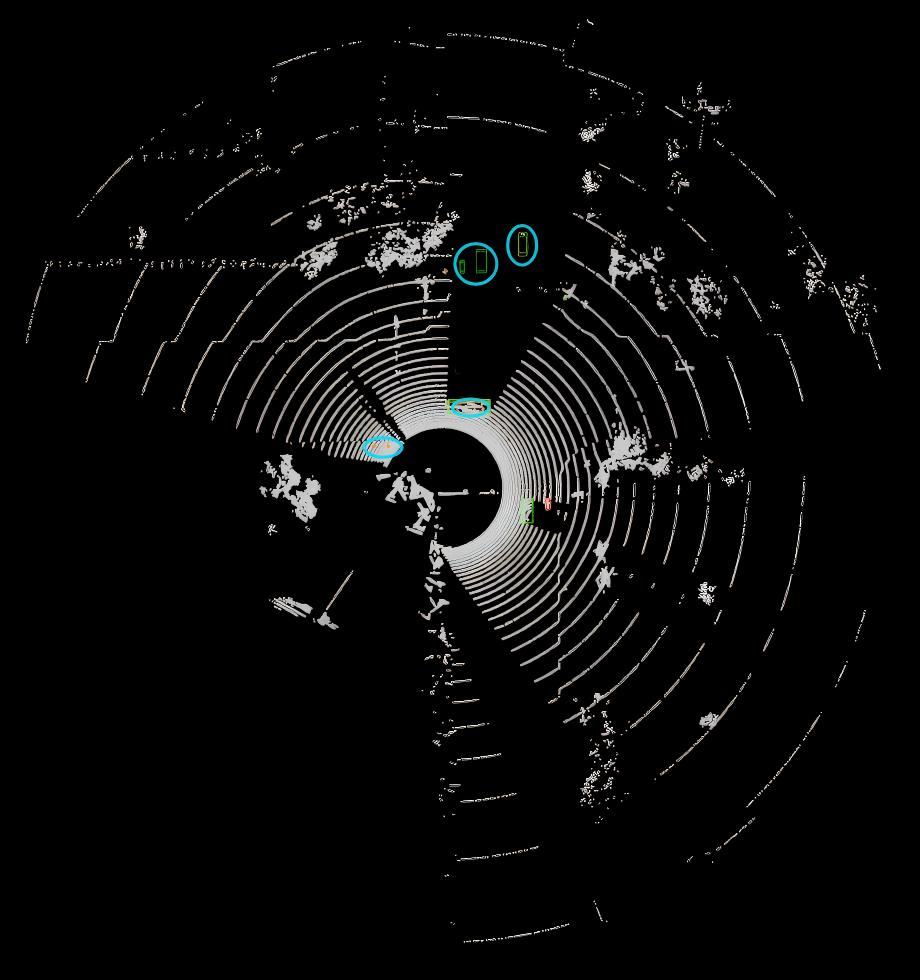} \\[4pt]

    \rotatebox{90}{CenterPoint} &
      \includegraphics[width=0.3\textwidth,height=3.3cm]{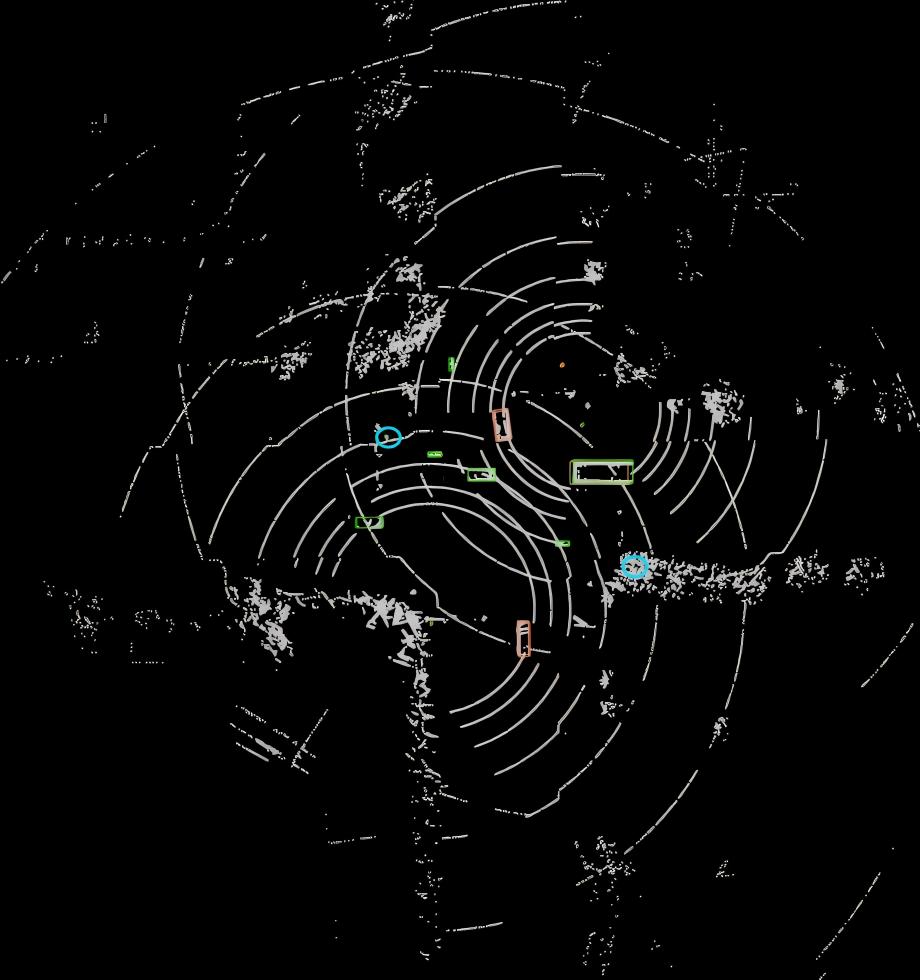} &
      \includegraphics[width=0.3\textwidth,height=3.3cm]{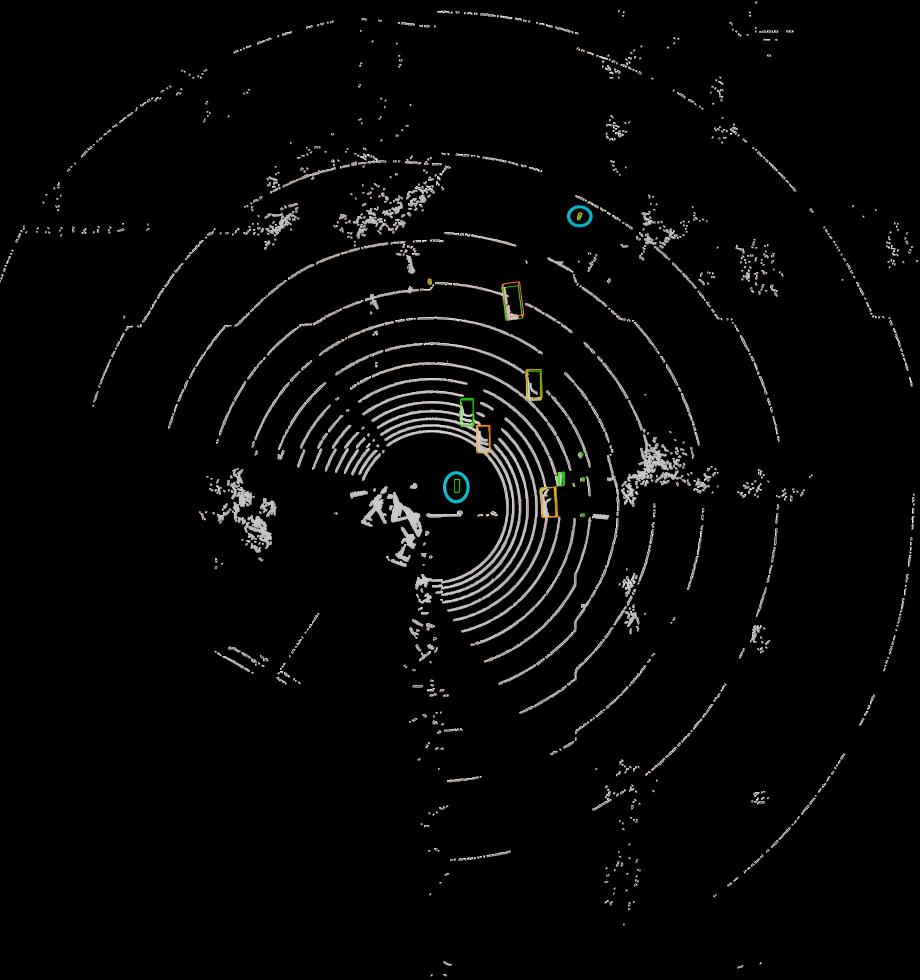} &
      \includegraphics[width=0.3\textwidth,height=3.3cm]{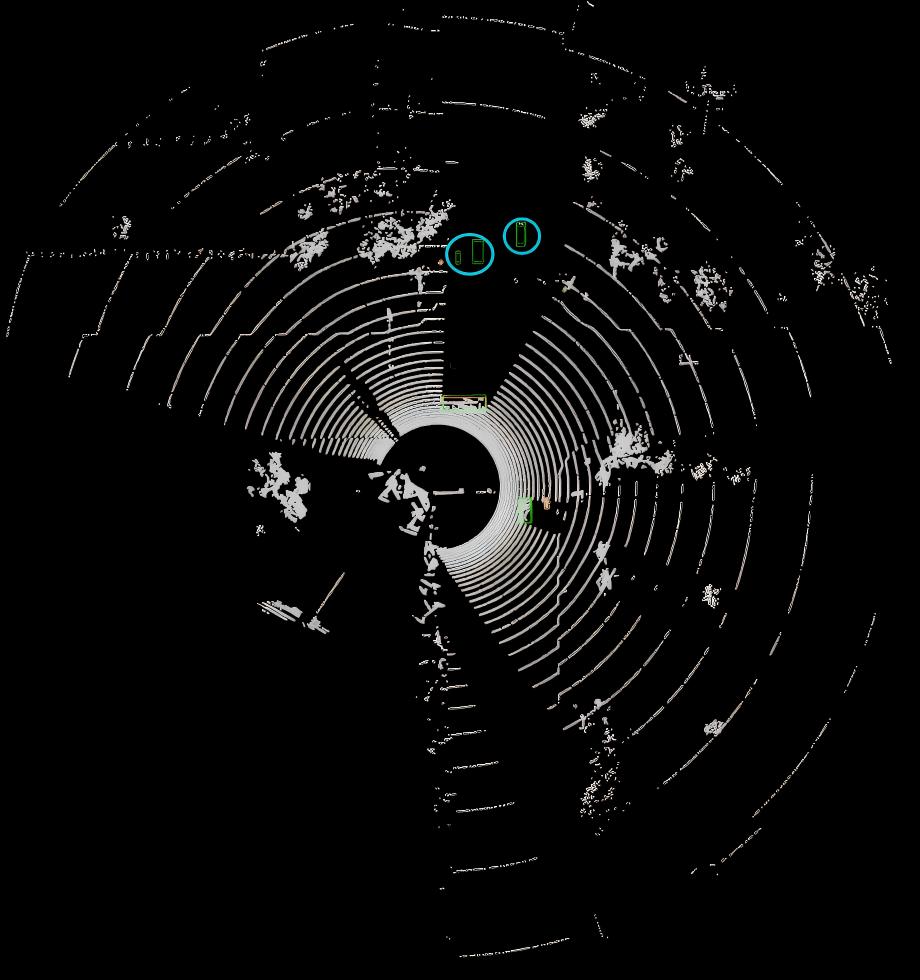} \\[4pt]

    \rotatebox{90}{PointPillars-LR} &
      \includegraphics[width=0.3\textwidth,height=3.3cm]{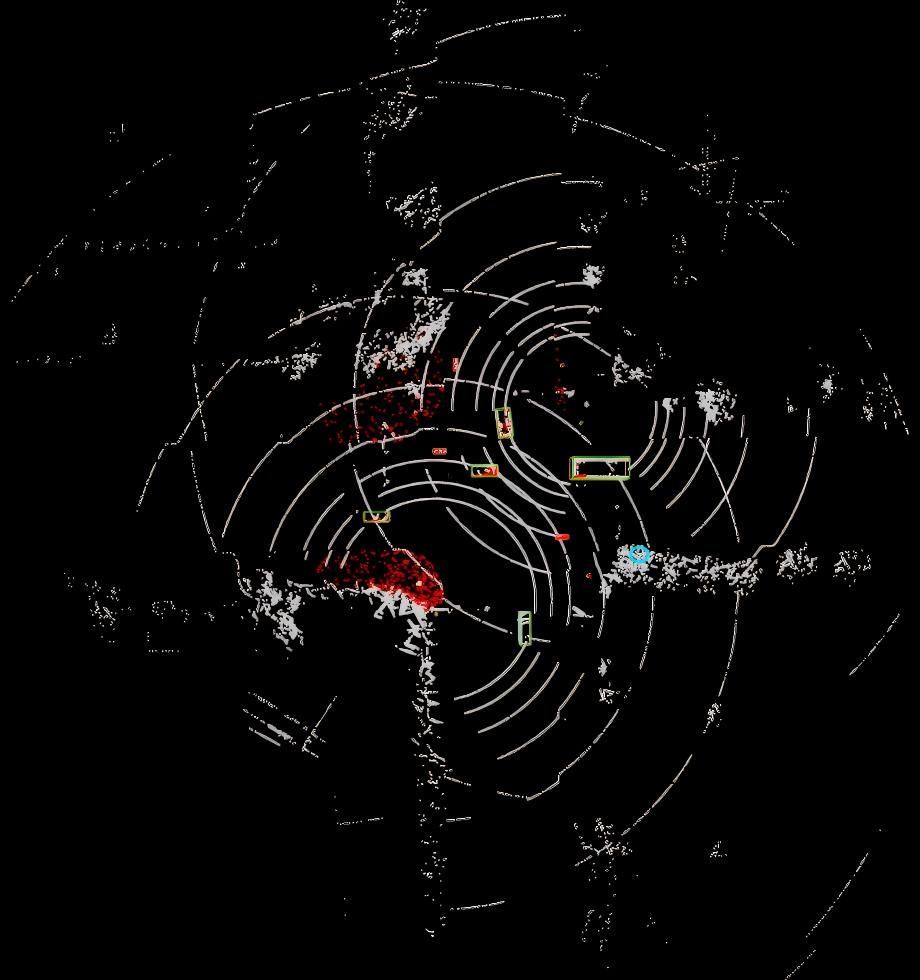} &
      \includegraphics[width=0.3\textwidth,height=3.3cm]{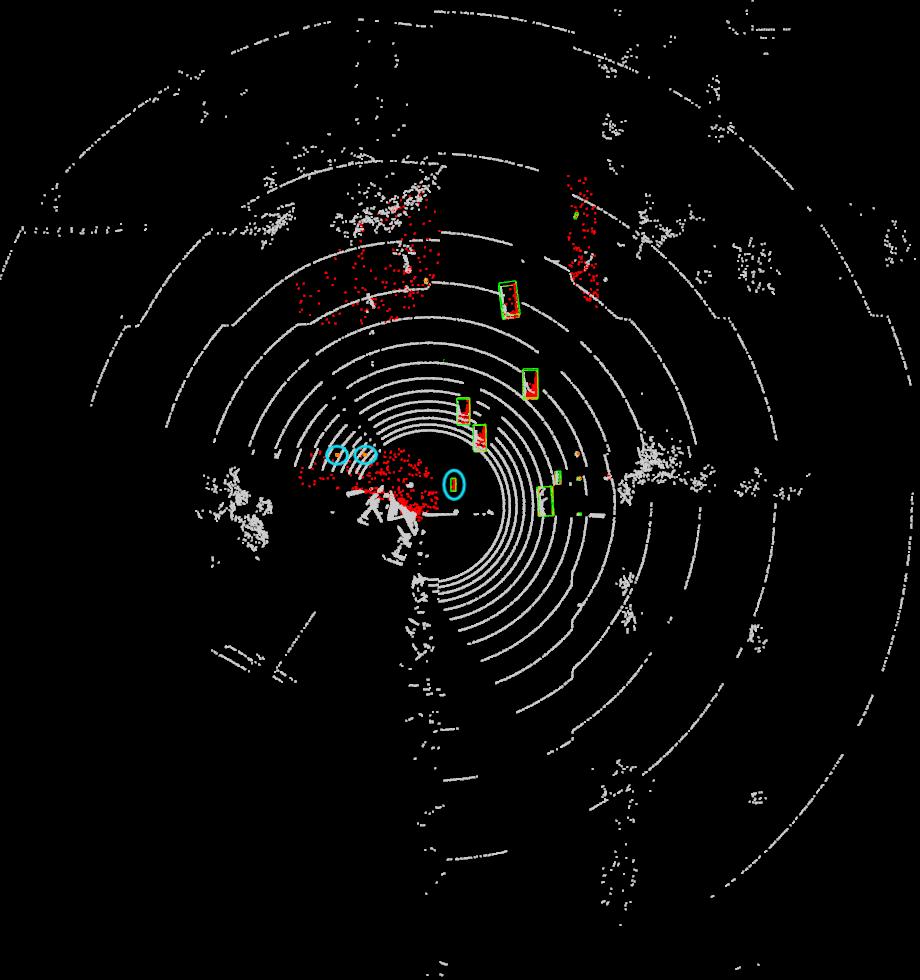} &
      \includegraphics[width=0.3\textwidth,height=3.3cm]{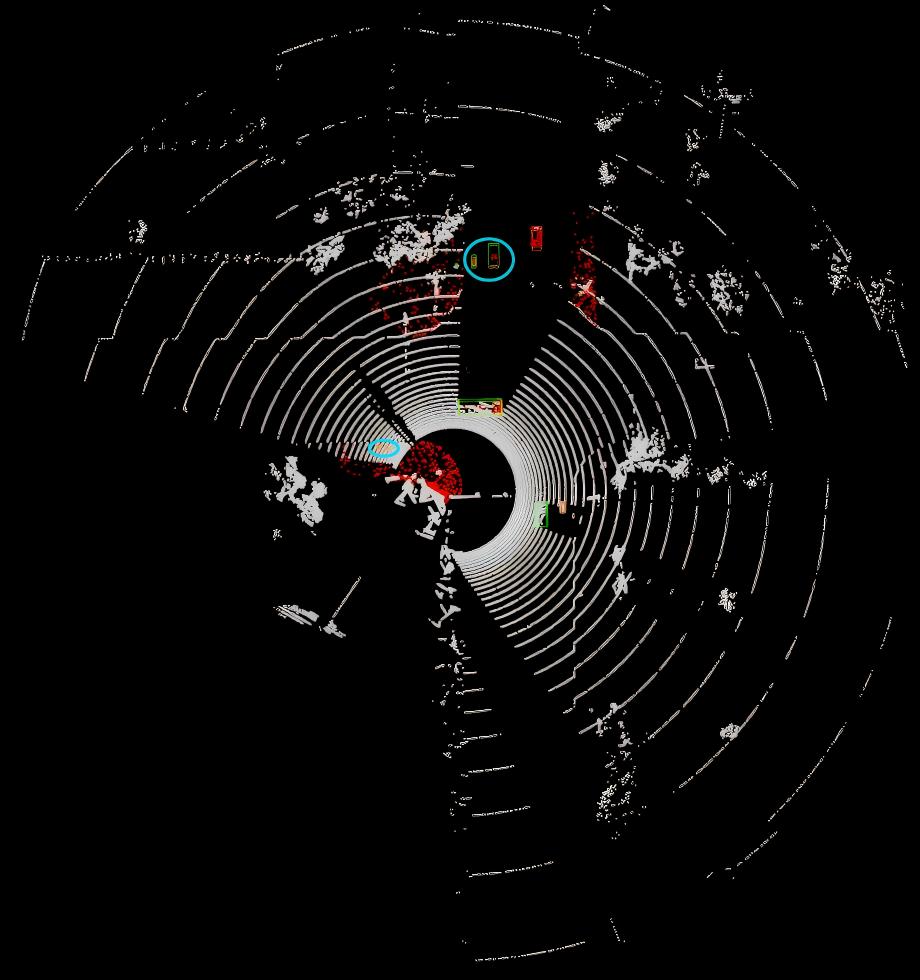} \\[4pt]
      
    \rotatebox{90}{L4DR} &
      \includegraphics[width=0.3\textwidth,height=3.3cm]{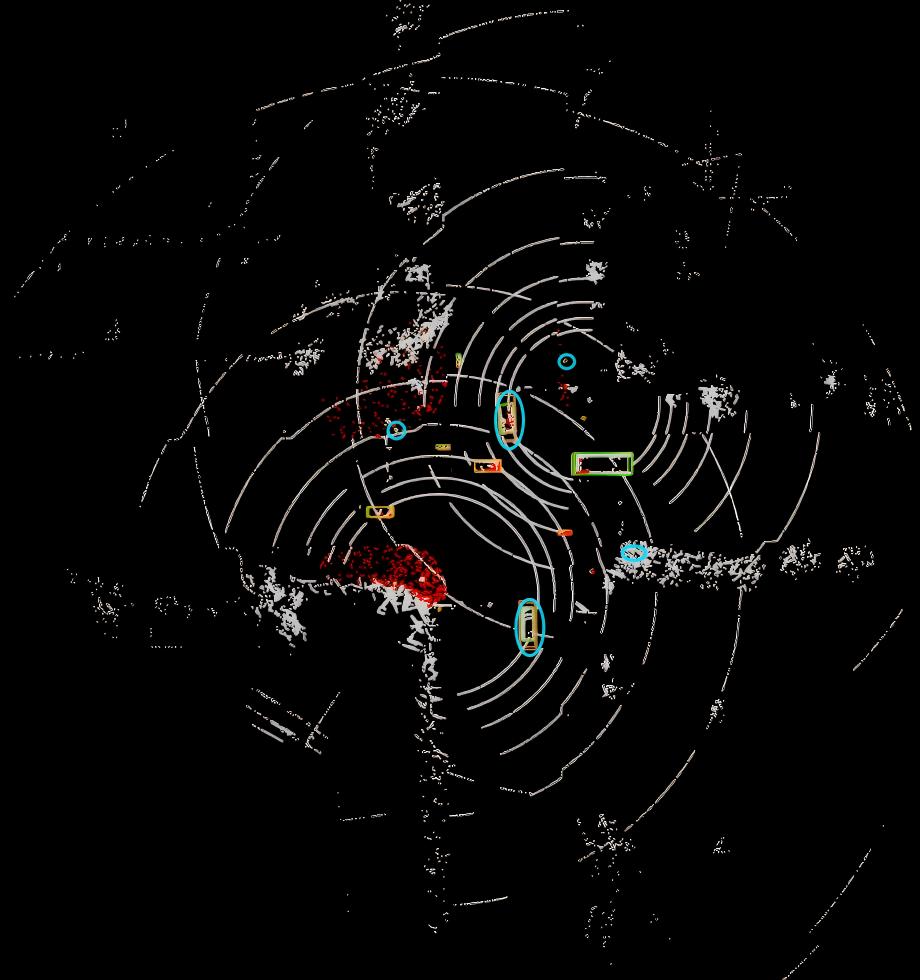} &
      \includegraphics[width=0.3\textwidth,height=3.3cm]{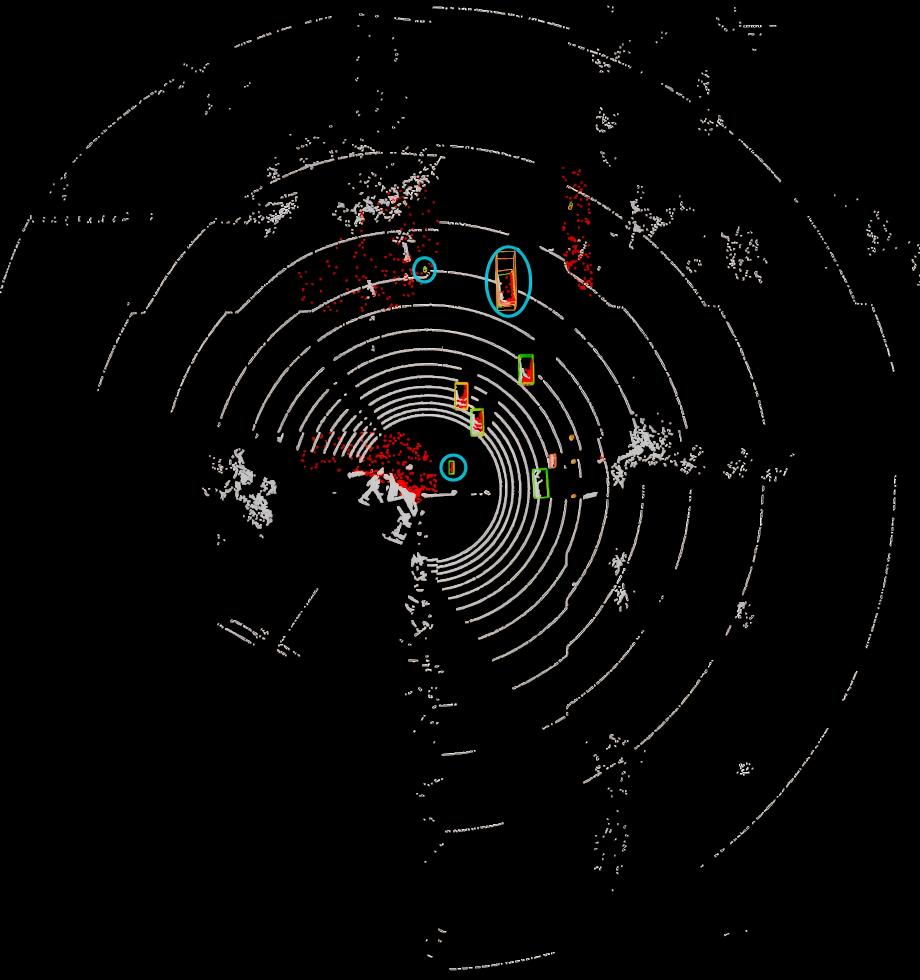} &
      \includegraphics[width=0.3\textwidth,height=3.3cm]{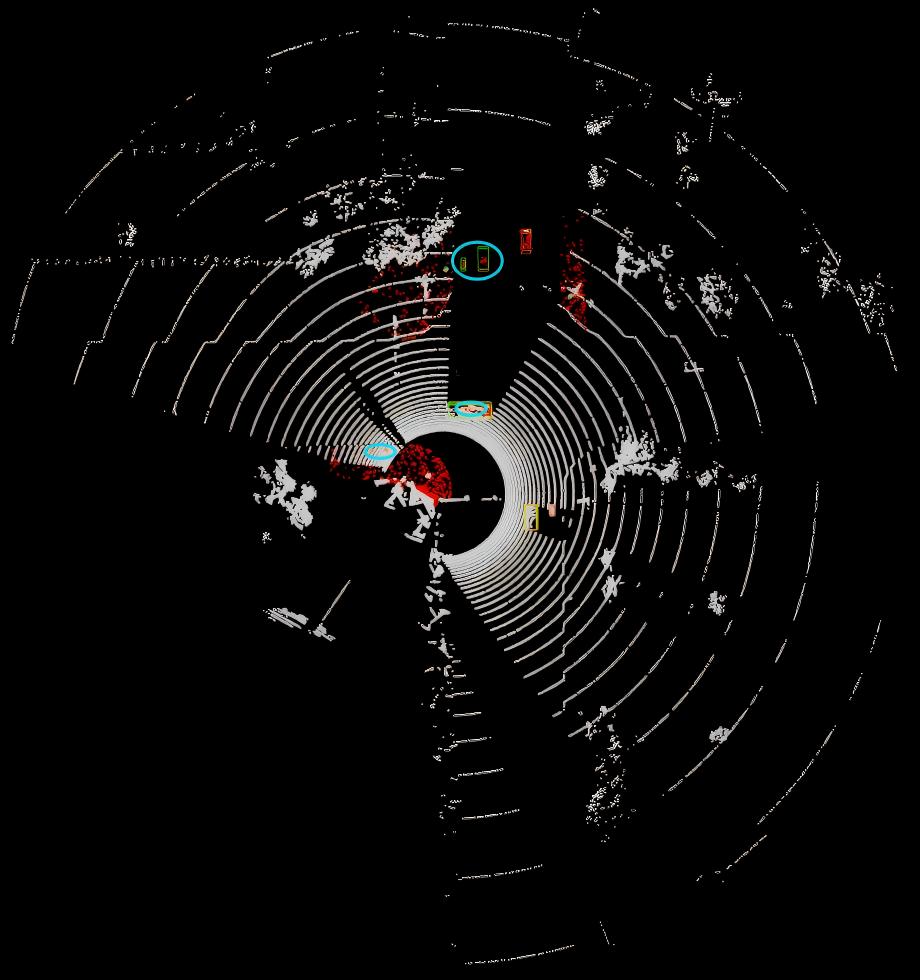} \\[4pt]

      \rotatebox{90}{LRFusion} &
      \includegraphics[width=0.3\textwidth,height=3.3cm]{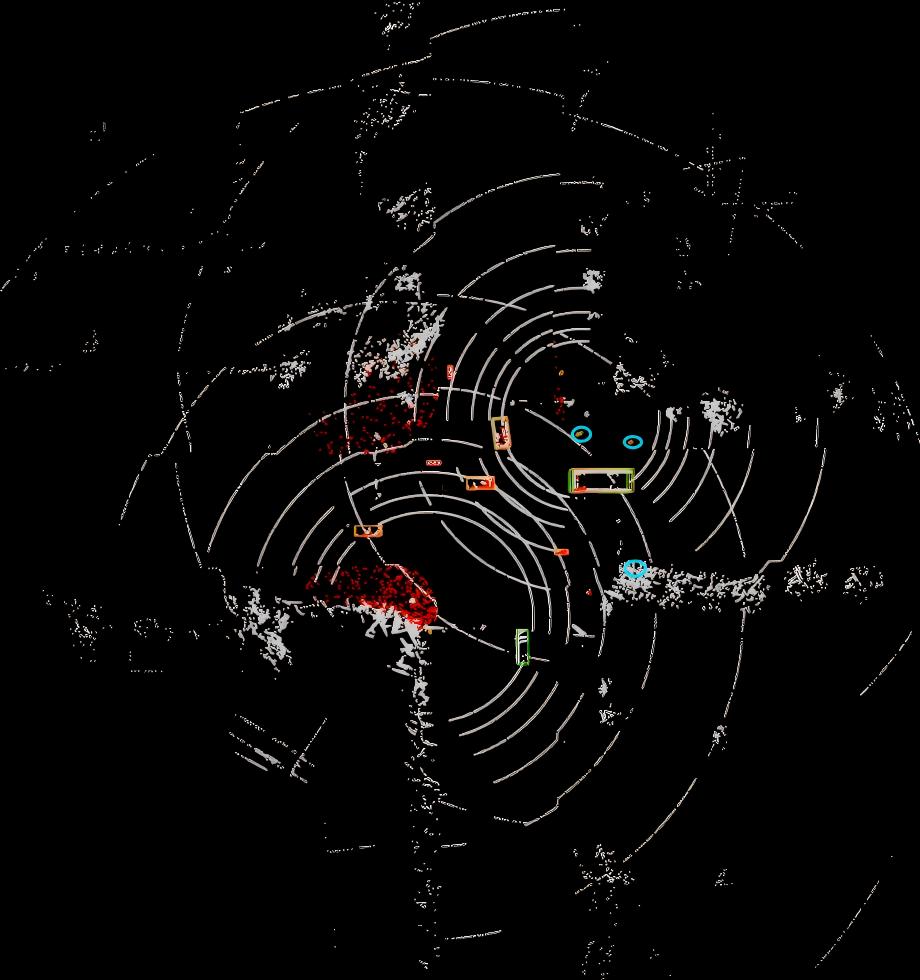} &
      \includegraphics[width=0.3\textwidth,height=3.3cm]{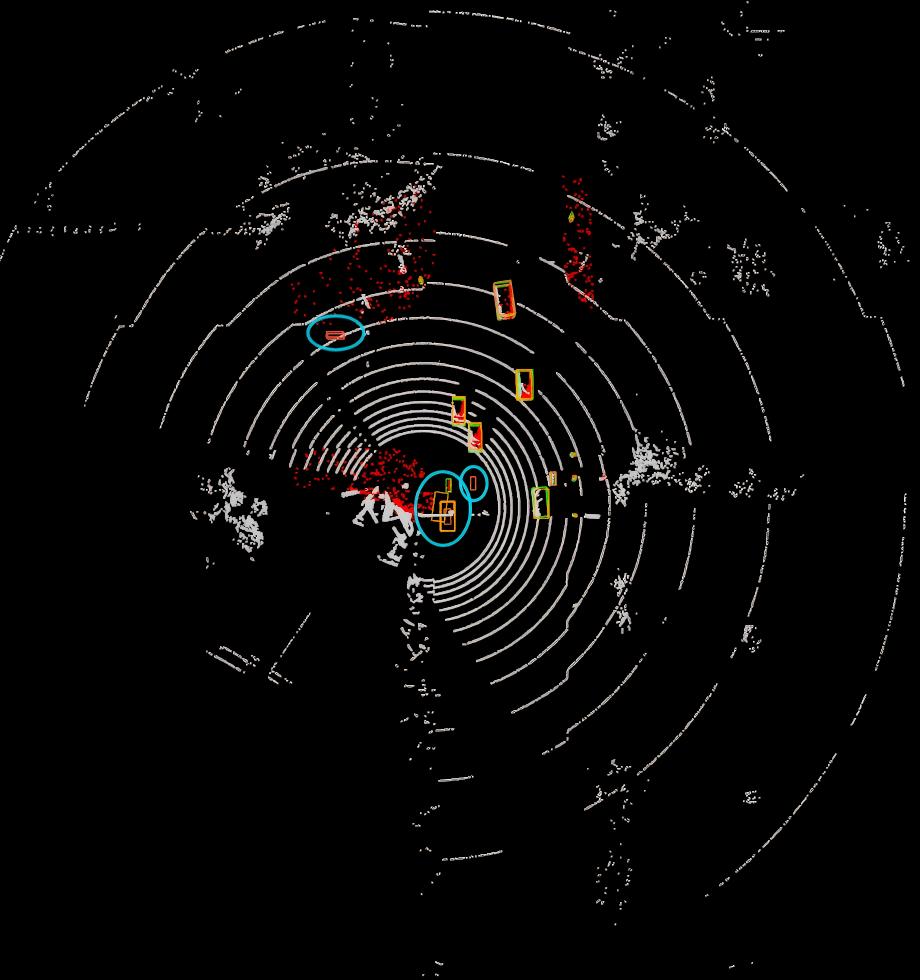} &
      \includegraphics[width=0.3\textwidth,height=3.3cm]{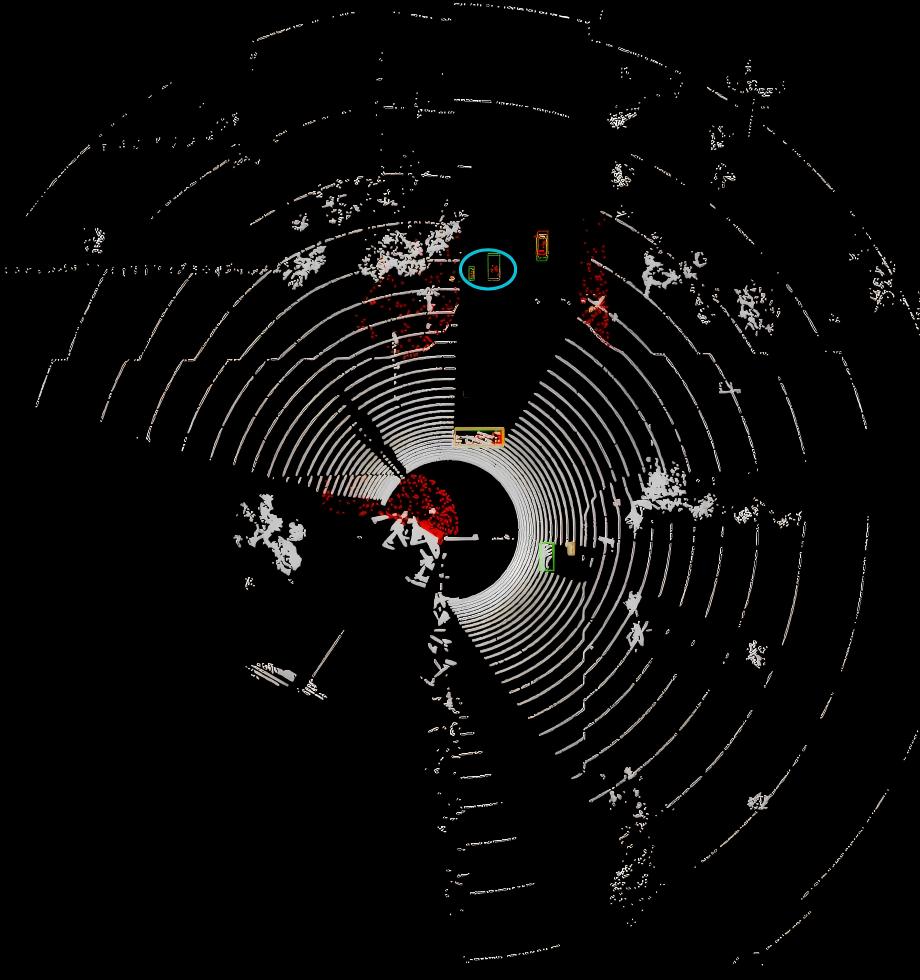} \\[4pt]

    \rotatebox{90}{Bi-lirafusion} &
      \includegraphics[width=0.3\textwidth,height=3.3cm]{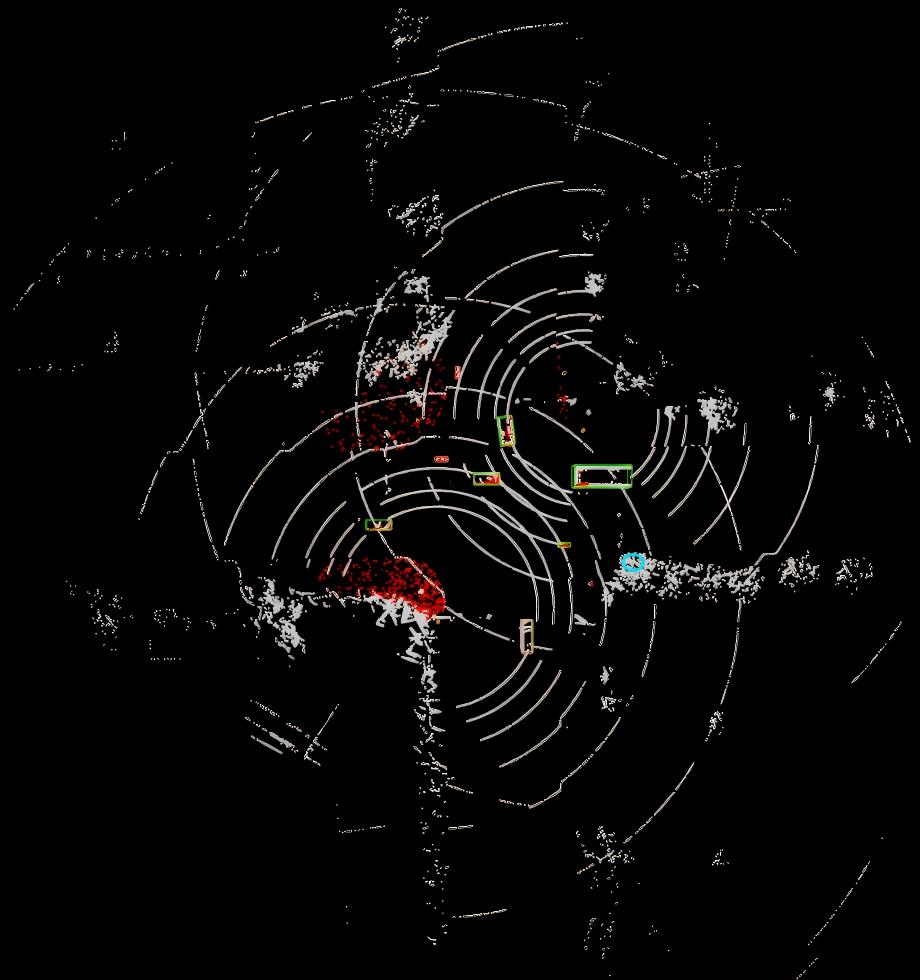} &
      \includegraphics[width=0.3\textwidth,height=3.3cm]{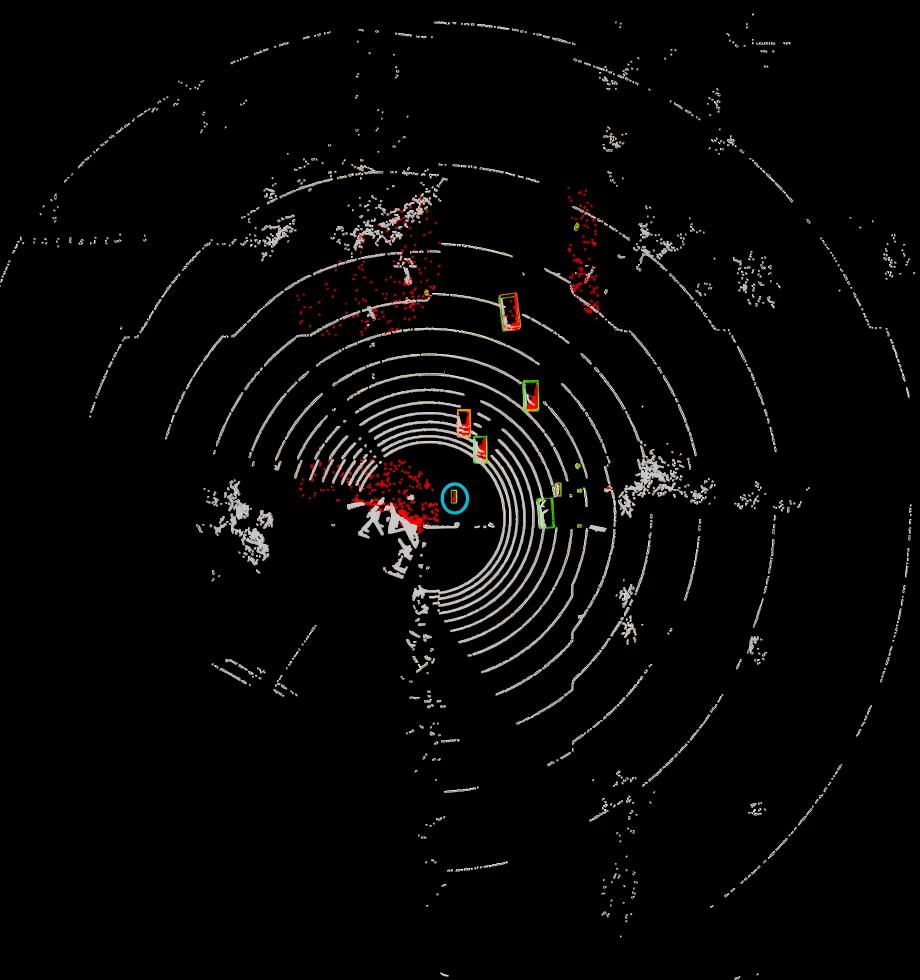} &
      \includegraphics[width=0.3\textwidth,height=3.3cm]{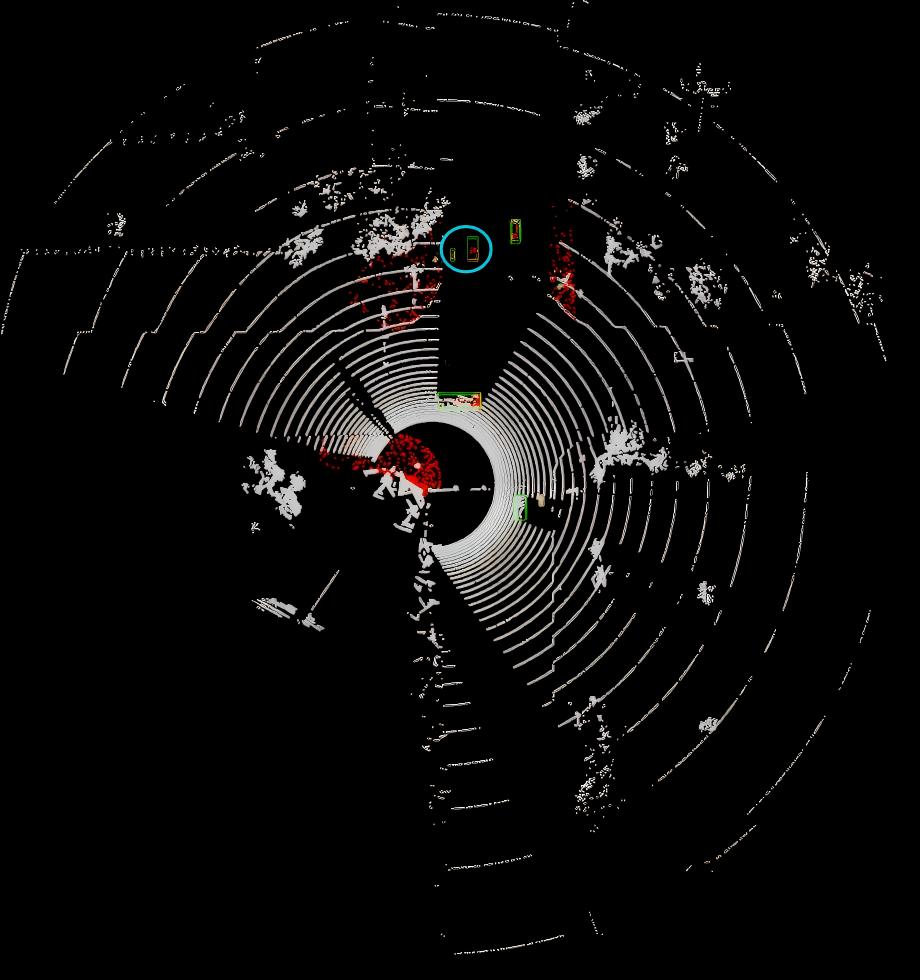} \\
      
  \end{tabular}
  \caption{Visualization performance comparison. We visualize 3D object detection results under three sensor configurations on our dataset with six perception algorithms(from up to bottom). The red and white points indicate 4D radar and LiDAR point clouds, respectively. We visualize the prediction and ground truth (green) boxes. Specifically, blue circles represent the missed or false detections of each algorithm.}
  \label{fig:vis_pred}
\end{figure*}

\clearpage

\bibliographystyle{IEEEtran}
\bibliography{ref}

\begin{thebibliography}{10}
\providecommand{\url}[1]{#1}
\csname url@samestyle\endcsname
\providecommand{\newblock}{\relax}
\providecommand{\bibinfo}[2]{#2}
\providecommand{\BIBentrySTDinterwordspacing}{\spaceskip=0pt\relax}
\providecommand{\BIBentryALTinterwordstretchfactor}{4}
\providecommand{\BIBentryALTinterwordspacing}{\spaceskip=\fontdimen2\font plus
\BIBentryALTinterwordstretchfactor\fontdimen3\font minus \fontdimen4\font\relax}
\providecommand{\BIBforeignlanguage}[2]{{%
\expandafter\ifx\csname l@#1\endcsname\relax
\typeout{** WARNING: IEEEtran.bst: No hyphenation pattern has been}%
\typeout{** loaded for the language `#1'. Using the pattern for}%
\typeout{** the default language instead.}%
\else
\language=\csname l@#1\endcsname
\fi
#2}}
\providecommand{\BIBdecl}{\relax}
\BIBdecl

\bibitem{bai2023cyber}
Z.~Bai, S.~P. Nayak, X.~Zhao, G.~Wu, M.~J. Barth, X.~Qi, Y.~Liu, E.~A. Sisbot, and K.~Oguchi, ``Cyber mobility mirror: A deep learning-based real-world object perception platform using roadside lidar,'' \emph{IEEE Transactions on Intelligent Transportation Systems}, vol.~24, no.~9, pp. 9476--9489, 2023.

\bibitem{zhao2019detection}
J.~Zhao, H.~Xu, H.~Liu, J.~Wu, Y.~Zheng, and D.~Wu, ``Detection and tracking of pedestrians and vehicles using roadside lidar sensors,'' \emph{Transportation research part C: emerging technologies}, vol. 100, pp. 68--87, 2019.

\bibitem{xu2023bridging}
R.~Xu, J.~Li, X.~Dong, H.~Yu, and J.~Ma, ``Bridging the domain gap for multi-agent perception,'' in \emph{2023 IEEE International Conference on Robotics and Automation (ICRA)}.\hskip 1em plus 0.5em minus 0.4em\relax IEEE, 2023, pp. 6035--6042.

\bibitem{xiang2024v2xreal}
\BIBentryALTinterwordspacing
H.~Xiang, Z.~Zheng, X.~Xia, R.~Xu, L.~Gao, Z.~Zhou, X.~Han, X.~Ji, M.~Li, Z.~Meng, L.~Jin, M.~Lei, Z.~Ma, Z.~He, H.~Ma, Y.~Yuan, Y.~Zhao, and J.~Ma, ``V2x-real: a largs-scale dataset for vehicle-to-everything cooperative perception,'' 2024. [Online]. Available: \url{https://arxiv.org/abs/2403.16034}
\BIBentrySTDinterwordspacing

\bibitem{lidarlibrary}
\BIBentryALTinterwordspacing
X.~Cai, W.~Jiang, R.~Xu, W.~Zhao, J.~Ma, S.~Liu, and Y.~Li, ``Analyzing infrastructure lidar placement with realistic lidar simulation library,'' in \emph{2023 IEEE International Conference on Robotics and Automation (ICRA)}.\hskip 1em plus 0.5em minus 0.4em\relax IEEE, May 2023. [Online]. Available: \url{http://dx.doi.org/10.1109/ICRA48891.2023.10161027}
\BIBentrySTDinterwordspacing

\bibitem{Qu2023}
A.~Qu, X.~Huang, and D.~Suo, ``Seip: Simulation-based design and evaluation of infrastructure-based collective perception,'' in \emph{2023 IEEE 26th International Conference on Intelligent Transportation Systems (ITSC)}, 2023, pp. 3871--3878.

\bibitem{jiang2023optimizing}
W.~Jiang, H.~Xiang, X.~Cai, R.~Xu, J.~Ma, Y.~Li, G.~H. Lee, and S.~Liu, ``Optimizing the placement of roadside lidars for autonomous driving,'' in \emph{Proceedings of the IEEE/CVF International Conference on Computer Vision}, 2023, pp. 18\,381--18\,390.

\bibitem{zou2022real}
Z.~Zou, R.~Zhang, S.~Shen, G.~Pandey, P.~Chakravarty, A.~Parchami, and H.~X. Liu, ``Real-time full-stack traffic scene perception for autonomous driving with roadside cameras,'' in \emph{2022 International Conference on Robotics and Automation (ICRA)}.\hskip 1em plus 0.5em minus 0.4em\relax IEEE, 2022, pp. 890--896.

\bibitem{Chen2024}
J.-K. Chen, M.-C. Lee, P.-C. Kang, and T.-S. Lee, ``Roadside radar network deployment and parameter optimization in road environments,'' \emph{IEEE Transactions on Vehicular Technology}, vol.~73, no.~8, pp. 11\,878--11\,894, 2024.

\bibitem{ma2024virtual}
Y.~Ma, Y.~Zheng, S.~Wang, Y.~D. Wong, and S.~M. Easa, ``Virtual-real-fusion simulation framework for evaluating and optimizing small-spatial-scale placement of cooperative roadside sensing units,'' \emph{Computer-Aided Civil and Infrastructure Engineering}, vol.~39, no.~5, pp. 707--730, 2024.

\bibitem{Vijay}
R.~Vijay, J.~Cherian, R.~Riah, N.~De~Boer, and A.~Choudhury, ``Optimal placement of roadside infrastructure sensors towards safer autonomous vehicle deployments,'' in \emph{2021 IEEE International Intelligent Transportation Systems Conference (ITSC)}, 2021, pp. 2589--2595.

\bibitem{lang2019pointpillars}
A.~H. Lang, S.~Vora, H.~Caesar, L.~Zhou, J.~Yang, and O.~Beijbom, ``Pointpillars: Fast encoders for object detection from point clouds,'' in \emph{Proceedings of the IEEE/CVF conference on computer vision and pattern recognition}, 2019, pp. 12\,697--12\,705.

\bibitem{yin2021centerbased3dobjectdetection}
\BIBentryALTinterwordspacing
T.~Yin, X.~Zhou, and P.~Krähenbühl, ``Center-based 3d object detection and tracking,'' 2021. [Online]. Available: \url{https://arxiv.org/abs/2006.11275}
\BIBentrySTDinterwordspacing

\bibitem{yang2020radarnetexploitingradarrobust}
\BIBentryALTinterwordspacing
B.~Yang, R.~Guo, M.~Liang, S.~Casas, and R.~Urtasun, ``Radarnet: Exploiting radar for robust perception of dynamic objects,'' 2020. [Online]. Available: \url{https://arxiv.org/abs/2007.14366}
\BIBentrySTDinterwordspacing

\bibitem{wang2023bilrfusionbidirectionallidarradarfusion}
\BIBentryALTinterwordspacing
Y.~Wang, J.~Deng, Y.~Li, J.~Hu, C.~Liu, Y.~Zhang, J.~Ji, W.~Ouyang, and Y.~Zhang, ``Bi-lrfusion: Bi-directional lidar-radar fusion for 3d dynamic object detection,'' 2023. [Online]. Available: \url{https://arxiv.org/abs/2306.01438}
\BIBentrySTDinterwordspacing

\bibitem{song2024lirafusiondeepadaptivelidarradar}
\BIBentryALTinterwordspacing
J.~Song, L.~Zhao, and K.~A. Skinner, ``Lirafusion: Deep adaptive lidar-radar fusion for 3d object detection,'' 2024. [Online]. Available: \url{https://arxiv.org/abs/2402.11735}
\BIBentrySTDinterwordspacing

\bibitem{huang2025l4drlidar4dradarfusionweatherrobust}
\BIBentryALTinterwordspacing
X.~Huang, Z.~Xu, H.~Wu, J.~Wang, Q.~Xia, Y.~Xia, J.~Li, K.~Gao, C.~Wen, and C.~Wang, ``L4dr: Lidar-4dradar fusion for weather-robust 3d object detection,'' 2025. [Online]. Available: \url{https://arxiv.org/abs/2408.03677}
\BIBentrySTDinterwordspacing

\bibitem{meredith1983interactions}
M.~A. Meredith and B.~E. Stein, ``Interactions among converging sensory inputs in the superior colliculus,'' \emph{Science}, vol. 221, no. 4608, pp. 389--391, 1983.

\bibitem{cappe2009multisensory}
C.~Cappe, E.~M. Rouiller, and P.~Barone, ``Multisensory anatomical pathways,'' \emph{Hearing research}, vol. 258, no. 1-2, pp. 28--36, 2009.

\bibitem{beauchamp2004unraveling}
M.~S. Beauchamp, B.~D. Argall, J.~Bodurka, J.~H. Duyn, and A.~Martin, ``Unraveling multisensory integration: patchy organization within human sts multisensory cortex,'' \emph{Nature neuroscience}, vol.~7, no.~11, pp. 1190--1192, 2004.

\bibitem{romanski2007representation}
L.~M. Romanski, ``Representation and integration of auditory and visual stimuli in the primate ventral lateral prefrontal cortex,'' \emph{Cerebral Cortex}, vol.~17, no. suppl\_1, pp. i61--i69, 2007.

\bibitem{2019Zhan}
W.~Zhan, L.~Sun, D.~Wang, Y.~Jin, and M.~Tomizuka, ``Constructing a highly interactive vehicle motion dataset,'' in \emph{2019 IEEE/RSJ International Conference on Intelligent Robots and Systems (IROS)}, 2019, pp. 6415--6420.

\bibitem{Zheng_2023}
\BIBentryALTinterwordspacing
O.~Zheng, M.~Abdel-Aty, L.~Yue, A.~Abdelraouf, Z.~Wang, and N.~Mahmoud, ``Citysim: A drone-based vehicle trajectory dataset for safety-oriented research and digital twins,'' \emph{Transportation Research Record: Journal of the Transportation Research Board}, vol. 2678, no.~4, p. 606–621, Jul. 2023. [Online]. Available: \url{http://dx.doi.org/10.1177/03611981231185768}
\BIBentrySTDinterwordspacing

\bibitem{dosovitskiy2017carla}
A.~Dosovitskiy, G.~Ros, F.~Codevilla, A.~Lopez, and V.~Koltun, ``Carla: An open urban driving simulator,'' in \emph{Conference on robot learning}.\hskip 1em plus 0.5em minus 0.4em\relax PMLR, 2017, pp. 1--16.

\bibitem{nuscenesdataset}
\BIBentryALTinterwordspacing
H.~Caesar, V.~Bankiti, A.~H. Lang, S.~Vora, V.~E. Liong, Q.~Xu, A.~Krishnan, Y.~Pan, G.~Baldan, and O.~Beijbom, ``nuscenes: A multimodal dataset for autonomous driving,'' 2020. [Online]. Available: \url{https://arxiv.org/abs/1903.11027}
\BIBentrySTDinterwordspacing

\end{thebibliography}

\end{document}